\pdfoutput=1

\documentclass[11pt]{article}

\usepackage[preprint]{acl}
\usepackage{float}
\usepackage{marvosym}
\usepackage{subcaption}

\usepackage{times}
\usepackage{latexsym}

\usepackage[T1]{fontenc}

\usepackage[utf8]{inputenc}

\usepackage{microtype}

\usepackage{inconsolata}

\usepackage{graphicx}
\usepackage[most]{tcolorbox}
\usepackage{lipsum} 
\usepackage{anyfontsize} 
\usepackage{listings}
\title{\textsc{Tower+}: \\Bridging Generality and Translation Specialization in Multilingual LLMs}

\author{Ricardo Rei\thanks{Core contributor. \Letter \, \url{ai-research@unbabel.com}}$^{1}$,~~Nuno M. Guerreiro$^{*1,2,3,4}$,~~José Pombal$^{*1,2,3}$,~~João Alves$^{*1}$\\
\bf Pedro Teixeirinha$^{1}$,~~Amin Farajian$^{1}$,~~André F. T. Martins$^{1,2,3}$
\\
$^{1}$Unbabel \\ $^{2}$Instituto de Telecomunicações \,\ \\
$^{3}$Instituto Superior Técnico \& Universidade de Lisboa (Lisbon ELLIS Unit) \\
$^{4}$MICS,
CentraleSupélec, Université Paris-Saclay
}

\usepackage{pgfplots}
\usepackage{pgfplotstable}
\usetikzlibrary{pgfplots.groupplots}
\pgfplotsset{compat=1.3}
\usepackage{tikz}
\usepackage{tikz-layers}
\usepackage{longtable}
\usetikzlibrary{patterns}
\usetikzlibrary{shapes.geometric} 
 \usetikzlibrary{arrows.meta}

\usepackage{CJKutf8}

\usepackage{hyperref}
\usepackage{url}

\usepackage{amssymb}
\usepackage{amsmath}
\usepackage{amsthm}
\usepackage{amsfonts}
\usepackage{bm}
\usepackage{nicefrac}
\usepackage{dsfont}
\usepackage{fancyvrb}
\usepackage{fvextra}
\usepackage{listings}
\usepackage{xcolor}
\usepackage{tabularx}
\usepackage[all]{nowidow}
\usepackage{xcolor}
\usepackage{colortbl}

\usepackage{booktabs}
\usepackage{multirow}
\usepackage{makecell}
\usepackage{arydshln}
\usepackage{rotating}  

\usepackage{comment}
\usepackage{xspace}
\usepackage{soul}

\usepackage{enumitem}
\usepackage[normalem]{ulem}
\setlist[itemize,enumerate]{leftmargin=*}

\usepackage{longtable}
\usepackage{colortbl}
\usepackage{xcolor}
\usepackage{graphicx}  
\usepackage{caption}
\definecolor{avg7color}{RGB}{0,112,192}    
\definecolor{avg15color}{RGB}{0,176,80}    
\definecolor{avgAllcolor}{RGB}{191,144,0}  

\makeatletter
\def\adl@drawiv#1#2#3{%
        \hskip.5\tabcolsep
        \xleaders#3{#2.5\@tempdimb #1{1}#2.5\@tempdimb}%
                #2\z@ plus1fil minus1fil\relax
        \hskip.5\tabcolsep}
\newcommand{\cdashlinelr}[1]{%
  \noalign{\vskip 2pt
           \global\let\@dashdrawstore\adl@draw
           \global\let\adl@draw\adl@drawiv}
  \cdashline{#1}[.4pt/2pt]
  \noalign{\global\let\adl@draw\@dashdrawstore
           \vskip 2pt}}
\makeatother

\definecolor{light-orange}{HTML}{fee9d4}
\definecolor{light-green}{HTML}{d8f0d3}
\definecolor{light-blue}{HTML}{dae8f5}
\definecolor{set10-red}{HTML}{e41a1c}
\definecolor{set10-blue}{HTML}{377eb8}
\definecolor{set10-green}{HTML}{4daf4a}

\definecolor{CustomBlue}{RGB}{57,83,191}
\definecolor{CustomRed}{HTML}{a75151}
\definecolor{BadRed}{HTML}{BF3C54}
\definecolor{GoodGreen}{HTML}{38761D}
\definecolor{DarkGreenOne}{RGB}{106,168,79}
\usepackage[most]{tcolorbox}

\newtcbox{\clustertab}[1]{on line, box align=base, colback={#1},colframe={#1},size=fbox,arc=2pt,top=-1.5pt, bottom=-1.5pt, left=-1.5pt, right=-1.5pt, boxrule=0pt, enlarge left by=1pt}

\newcommand{\tocite}[1]{{\textcolor{orange}{[CITE]}}}

\newcommand{\Towervtwo}{\textsc{Tower-v2}}
\newcommand{\Towerplus}{\textsc{Tower+}}
\newcommand{\Towerpluslarge}{\textsc{Tower+} 72B}
\newcommand{\Towerplusmedium}{\textsc{Tower+} 9B}
\newcommand{\Towerplussmall}{\textsc{Tower+} 2B}

\newcommand{\almar}{\textsc{Alma-R}}
\newcommand{\gemmax}{\textsc{GemmaX}}
\newcommand{\gemmatwoxl}{\textsc{Gemma-2}}
\newcommand{\gemmatwoxxl}{\textsc{Gemma-2}}
\newcommand{\llamathree}{\textsc{LLaMA-3.3}}
\newcommand{\qwenseventy}{\textsc{Qwen-2.5}}

\newcommand{\comet}{\textsc{Comet}}

\newcommand{\chrf}{\textsc{chrF}}

\newcommand{\xcomet}{\textsc{xComet-xxl}}
\newcommand{\gptfouro}{\textsc{GPT-4o-1120}}

\newcommand{\claudethreeseven}{\textsc{Claude-Sonnet-3.7}}

\newcommand{\wmttwofour}{\textsc{WMT24++}}

\newcommand{\metricx}{\textsc{MetricX24-xxl}}

\usepackage{pifont}
\begin{document}

\maketitle
\begin{abstract}
Fine-tuning pretrained LLMs has been shown to be an effective strategy for reaching state-of-the-art performance on specific tasks like machine translation.
However, this process of adaptation often implies sacrificing general-purpose capabilities, such as conversational reasoning and instruction-following, hampering the utility of the system in real-world applications that require a mixture of skills.
In this paper, we introduce \textsc{Tower+}, a suite of models designed to deliver strong performance across both translation and multilingual general-purpose text capabilities. We achieve a Pareto frontier between translation specialization and multilingual general-purpose capabilities by introducing a novel training recipe that builds on \textsc{Tower}~\citep{alves2024tower}, comprising continued pretraining, supervised fine-tuning, preference optimization, and reinforcement learning with verifiable rewards. At each stage of training, we carefully generate and curate data to strengthen performance on translation as well as general-purpose tasks involving code generation, mathematics problem solving, and general instruction-following.
We develop models at multiple scales: 2B, 9B, and 72B. Our smaller models often outperform larger general-purpose open-weight and proprietary LLMs (e.g., \textsc{Llama 3.3} 70B, GPT-4o). Our largest model delivers best-in-class translation performance for high-resource languages and top results in multilingual Arena Hard evaluations and in IF-MT, a benchmark we introduce for evaluating both translation and instruction-following. Our findings highlight that it is possible to rival frontier models in general capabilities, while optimizing for specific business domains, such as translation and localization.

\end{abstract}

\section{Introduction}

\definecolor{metablue}{HTML}{0064E0}
\definecolor{qwenpurple}{HTML}{6337e6}
\definecolor{goldenbrown}{rgb}{0.6, 0.4, 0.08}
\definecolor{battleshipgrey}{rgb}{0.3, 0.3, 0.3}

\begin{figure*}[t]
    \centering
    \pgfplotsset{
        footnotesize,
        samples=10,
        xmin=80, xmax=88,
        ymin=-5, ymax=76,
        xtick={80,81,82,83,84,85,86,87,88},
        xticklabels={80,81,82,83,84,85,86,87,88},
        ytick={0,10,20,30,40,50,60,70},
        yticklabels={0,10,20,30,40,50,60,70},
        grid style={dashed,color=gray!20},
        grid=both,
        xlabel=Translation Performance (\xcomet{} on \wmttwofour{}),
        x label style={at={(axis description cs:0.5,-0.11)},anchor=north},
        ylabel=General Capabilities Performance \\ (M-ArenaHard),
        y label style={at={(axis description cs:1,0.485)},anchor=south},
    }
    
    \begin{tikzpicture}
    \begin{groupplot}[
        group style = {group size = 1 by 1, horizontal sep = 24pt},
        width = 0.9\textwidth, 
        height = 5cm
    ]
        \nextgroupplot[
            align = center,
            legend style={
                at={(0.5,1.25)},
                anchor=north,
                legend columns=2,
                column sep=10pt,
                draw=none,
            },
            legend cell align={left},
            axis x line*=bottom,
            axis y line*=left,
            y label style={at={(axis description cs:-0.04,0.485)},anchor=south},
            xtick pos=bottom,
            ytick pos=left,
        ]
        \addplot[gray!50, line width=0.75pt, forget plot] coordinates {(81.93, 13.38) (86.25, 33.47)};
        \addplot[gray!50, line width=0.75pt, forget plot] coordinates {(84.72, 50.0) (86.68, 54.52)};
        \node at (axis cs:86.40, 4.01) [anchor=center] {\includegraphics[height=0.4cm]{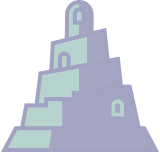}};
        \node[anchor=north, yshift=-2pt] at (axis cs:86.40, 4.01) {\tiny\shortstack{\textsc{Tower v2}}};
        \node at (axis cs:82.74, 13.15) [anchor=center] {\includegraphics[height=0.25cm]{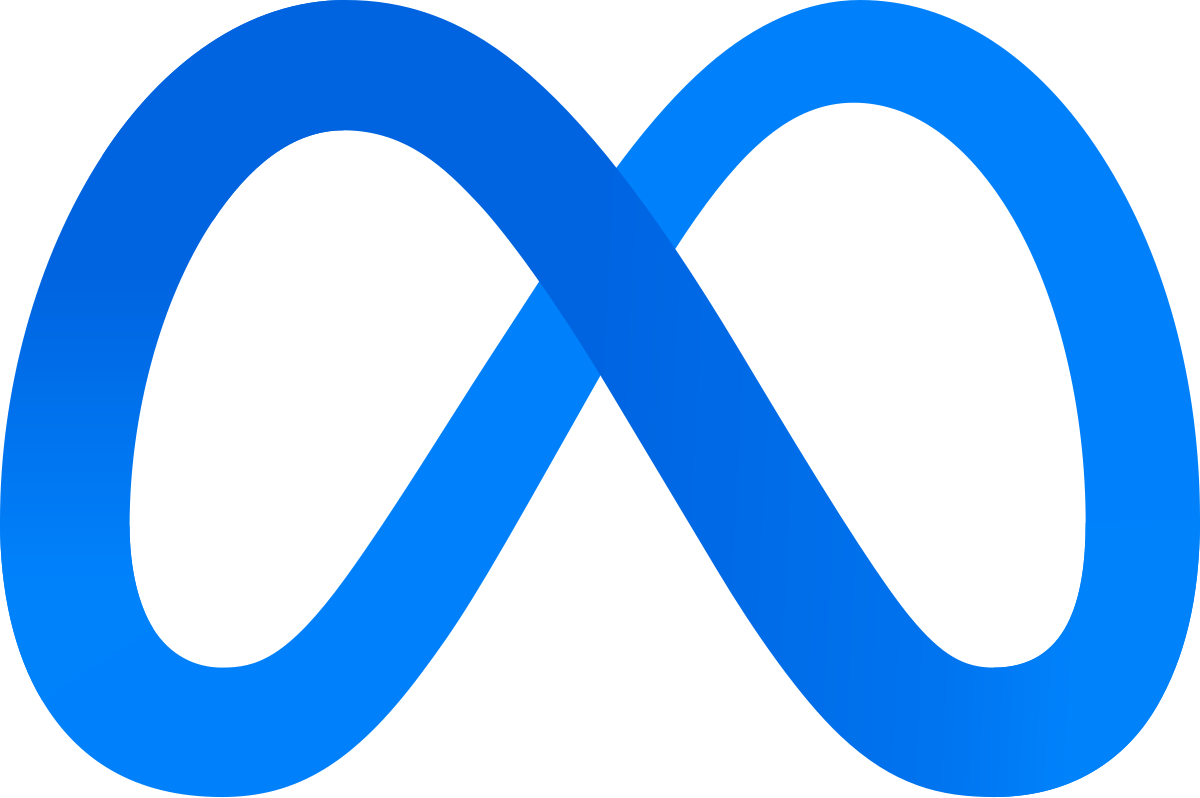}};
        \node[anchor=north, yshift=-2pt, xshift=5pt] at (axis cs:82.74, 13.15) {\tiny\shortstack{\textsc{Llama 3.3 70B}}};
        \node at (axis cs:84.72, 50.0) [anchor=center] {\includegraphics[height=0.4cm]{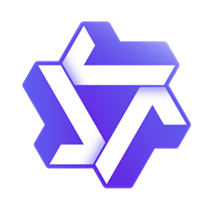}};
        \node[anchor=north, yshift=-2pt] at (axis cs:84.72, 50.0) {\tiny\shortstack{\textsc{Qwen 2.5 72B}}};
        \node at (axis cs:81.93, 13.38) [anchor=center] {\includegraphics[height=0.4cm]{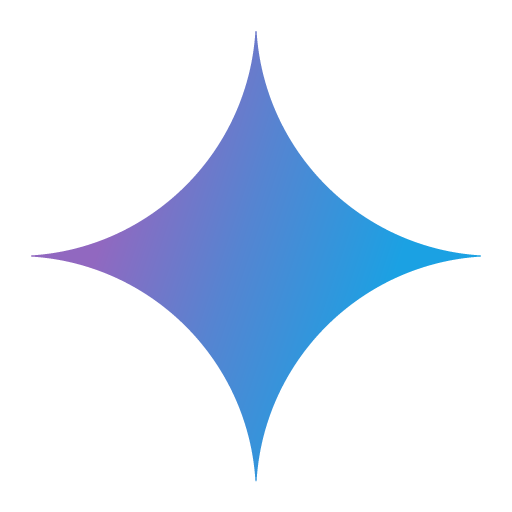}};
        \node[anchor=north, yshift=14pt] at (axis cs:81.93, 13.38) {\tiny\shortstack{\textsc{Gemma 2 9B}}};
        \node at (axis cs:83.68, 22.81) [anchor=center] {\includegraphics[height=0.4cm]{latex/logos/gemini.png}};
        \node[anchor=north, yshift=-2pt] at (axis cs:83.68, 22.81) {\tiny\shortstack{\textsc{Gemma 2 27B}}};
        \addplot[color=goldenbrown!50, line width=1.1pt, mark=*, mark size=3.5pt, draw=goldenbrown!60] coordinates {(84.26, 0.02)};
        \node[anchor=north, yshift=14pt,xshift=10pt] at (axis cs:84.26, 0.02) {\tiny\shortstack{\textsc{GemmaX 9B}}};
        \addplot[color=battleshipgrey!50, line width=1.1pt, mark=*, mark size=3.5pt, draw=battleshipgrey!60] coordinates {(80.97, 0.2)};
        \node[anchor=north, yshift=15pt,xshift=0pt] at (axis cs:80.97, 0.2) {\tiny\shortstack{\textsc{ALMA-R 13B}}};
        \node at (axis cs:86.69, 61.19) [anchor=center] {\includegraphics[height=0.3cm]{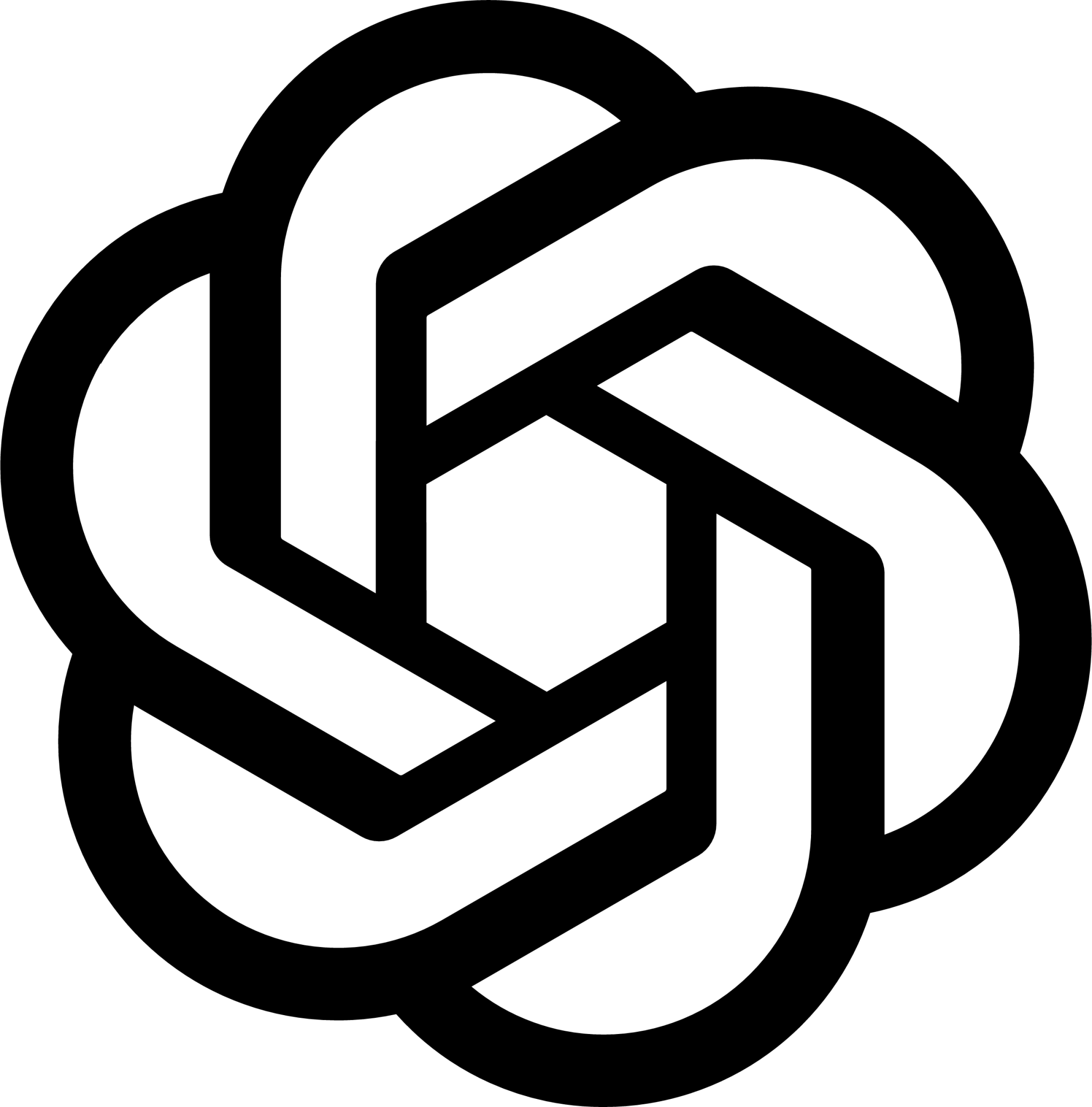}};
        \node[anchor=north, yshift=10pt,xshift=15pt] at (axis cs:86.69, 61.19) {\tiny\shortstack{\textsc{GPT-4o}}};
        \node at (axis cs:86.41, 67.00) [anchor=center] {\includegraphics[height=0.4cm]{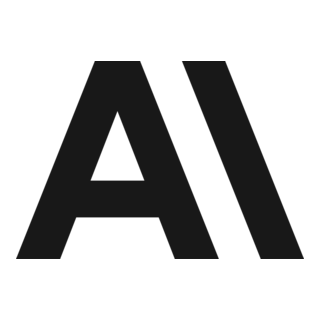}};
        \node[anchor=north, yshift=14pt] at (axis cs:86.41, 67.00) {\tiny\shortstack{\textsc{Claude Sonnet 3.7}}};
        \node at (axis cs:86.25, 33.47) [anchor=center] {\includegraphics[height=0.4cm]{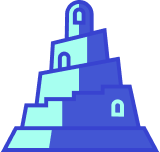}};
        \node[anchor=north, yshift=-3pt] at (axis cs:86.25, 33.47) {\tiny\shortstack{\textsc{\textbf{\Towerplusmedium{}}}}};
        \node at (axis cs:86.68, 54.52) [anchor=center] {\includegraphics[height=0.4cm]{latex/logos/tower.png}};
        \node[anchor=north, yshift=-3pt] at (axis cs:86.68, 54.52) {\tiny\shortstack{\textsc{\textbf{\Towerpluslarge{}}}}};
    \end{groupplot}
    \end{tikzpicture}
\caption{Translation and general capabilities performance of state-of-the-art translation-specific (circles) and general-purpose LLMs (logos) of varying sizes (from 9B to 72B). For their size, \textsc{Tower+} models outperform or match open-weight models on both axes, and \textsc{Tower+} 72B is competitive to state-of-the-art closed models. We omit 2B models for better visualization; we show detailed results in Table~\ref{tab:main_results}. Gray lines connect each \textsc{Tower+} model to their respective ``instruction-tuned'' counterpart.}
\label{fig:pareto_frontier}
\end{figure*}

Large Language Models (LLMs) are rapidly emerging as the \textit{de facto} solution for multilingual machine translation. Recent studies and shared tasks have shown that state-of-the-art proprietary LLMs, such as GPT-4o and Claude 3.7 are currently state-of-the-art in translation quality~\cite{kocmi-etal-2023-findings, kocmi-etal-2024-findings, deutsch2025wmt24expandinglanguagecoverage}. 
At the same time, several studies have shown that open-weight LLMs can be effectively adapted for machine translation, reaching parity with or even surpassing the translation quality of leading proprietary models~\cite{alves2024tower, rei-etal-2024-tower, xu2025xalma, gemmax2}. However, these task-specific adaptations often come at the cost of general-purpose capabilities\footnote{The term ``general-purpose capabilities'' refers to the real-world utility of language models in handling queries that require core knowledge, instruction-following, and conversational reasoning~\citep{zheng2023judging}} such as instruction following and conversational reasoning. For example, \textsc{Tower v2}, the best-performing system in the WMT24 shared task, ranked first in 8 out of 11 language pairs~\cite{kocmi-etal-2024-findings}, yet underperforms models like GPT-4o or Claude 3 on general chat evaluations. This tradeoff is illustrated in Figure~\ref{fig:pareto_frontier}, which shows that translation-specialized models tend to fall short on the Pareto frontier that balances translation quality and general-purpose utility. In addition, the degradation of instruction-following capabilities can hinder the ability to handle complex translation scenarios, such as adhering to a terminology or formatting rules.
Indeed, translation-specific systems greatly underperform general-purpose models on IF-MT, a novel benchmark we introduce for evaluating both translation and instruction-following.

To address this challenge, we investigate how to develop state-of-the-art translation models without compromising performance on general chat benchmarks. Our post-training pipeline builds on the framework introduced in earlier \textsc{Tower} models~\cite{alves2024tower, rei-etal-2024-tower}, but introduces several important methodological improvements.

We begin with continued pretraining (CPT) on a curated mixture of monolingual and parallel data to enhance multilingual fluency and translation performance. Unlike previous versions, the CPT stage now also includes a small fraction (1\%) of high-quality instruction-like data, which helps preserve general capabilities during early specialization. For supervised fine-tuning (SFT), we refine the data mixture and significantly rebalance task weights: while translation remains a key component, it now accounts for only 22\% of the SFT corpus, with the remainder covering instruction-following tasks such as mathematics, code generation, and question answering. In addition, we improve the quality of SFT responses through automated refinement using open-weight models, resulting in better quality answers.

Finally, we introduce a new preference optimization stage using Weighted Preference Optimization~\citep[WPO]{zhou-etal-2024-wpo}, complemented by Group Relative Policy Optimization~\citep[GRPO]{shao2024deepseekmath} and verifiable reward models (RLVR), to further align the model with high-quality outputs across tasks. We evaluate our models on both general chat benchmarks and translation tasks. Crucially, we analyze the contribution of each stage of the training pipeline in Section~\ref{sec:impact_stages}, and assess the role of the base model in balancing general-purpose and translation-specific capabilities in Section~\ref{sec:base_models}. Our contributions are as follows:

\begin{itemize}
\item We present, to the best of our knowledge, the first systematic study on balancing translation quality and general-purpose capabilities in open-weight LLMs. While most prior work \citep{alves2024tower, xualmar, rei-etal-2024-tower, gemmax2} has focused solely on maximizing translation performance, our approach explicitly targets a broader trade-off.
\item We introduce a post-training pipeline that integrates diverse multilingual signals without compromising general chat abilities. Our approach can serve as a blueprint for adapting LLMs to domain- or task-specific business use cases while preserving general capabilities.
\item We introduce and release IF-MT,\footnote{We make IF-MT and all \Towerplus{} models available on \href{https://huggingface.co/collections/Unbabel/tower-plus-6846ca452a10c0905dc03c0f}{Huggingface}\label{ftn:releases}.} a novel benchmark for evaluating both translation and instruction-following capabilities on two langauge pairs (English$\rightarrow$Spanish and English$\rightarrow$Chinese).
\item We release \Towerplus{},\textsuperscript{\ref{ftn:releases}} a suite of models that demonstrate strong performance across translation, general capabilities, and a benchmark that mixes the two. We match or exceed the translation quality of prior \textsc{Tower} models and \gptfouro{}, while also surpassing general-purpose open-weight models like Llama-3.3 70B and Qwen2.5 72B on M-ArenaHard.
\end{itemize}

\begin{figure*}[h!]
    \centering
    \includegraphics[width=1.0\linewidth,trim={5.5cm 4.7cm 5.5cm 2.9cm},clip]{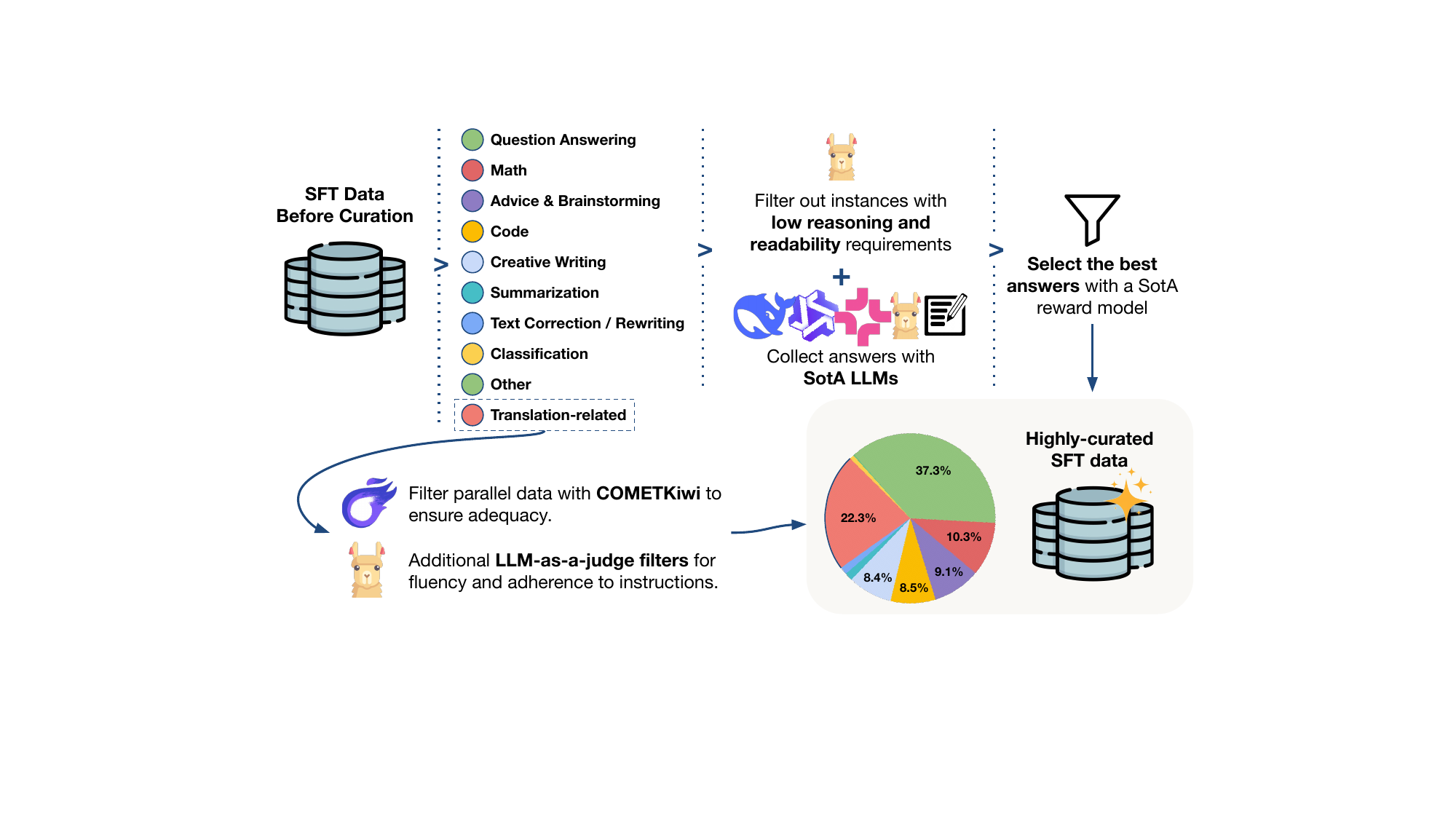}
    \caption{Process for creating and curating data for our final dataset for SFT.}
    \label{fig:sft-process}
\end{figure*}

\section{\textsc{Tower} Post-Training}

Our refined post-training pipeline consists of four stages: Continued Pretraining (CPT), supervised fine-tuning (SFT) (as in \Towervtwo{}, \citealt{rei-etal-2024-tower}), preference optimization using WPO~\cite{zhou-etal-2024-wpo}, and finally, reinforcement learning with verifiable rewards using GRPO~\cite{shao2024deepseekmath}. In this section, we describe each stage in detail and explain how translation signals are incorporated throughout the pipeline.

\subsection{Continued Pretraining}

The continued pretraining phase leverages monolingual, parallel, and general-purpose instruction-following data. Similarly to previous \textsc{Tower} models, the data distribution follows a 66\%/33\% split between monolingual and parallel data; in this version, we include 1\% of instruction-following data. 

All monolingual data is sourced from FineWeb-Edu~\citep{penedo2024fineweb}. Most parallel data is sourced from OPUS~\cite{TIEDEMANN12.463} and filtered using \textsc{CometKiwi}~\cite{rei-etal-2022-cometkiwi}. Additionally, the parallel data is formatted as a translation instruction followed by the corresponding translation.\footnote{We use multiple templates to prepare the parallel corpora. See examples in Appendix Figure \ref{fig:translation-prompts}.} For language pairs where it is available, we include document-level translation data from EuroParl~\citep{koehn2005europarl}, ParaDocs~\citep{wicks2024recovering}, and CosmoPedia-v2~\citep{benallal2024smollmcorpus}, each totaling 10\% of the data of each language pair. Instruction-following data is sampled from FineWeb-Edu using \texttt{dsir}~\citep{xie2023data} to be similar to high-quality instruction-following data.
For monolingual data, we apply the EuroFilter-v1~\citep{martins2025eurollm} quality filter, a multilingual educational classifier built using mDeBERTA~\cite{he2023debertav}.

Using these high-quality parallel and monolingual datasets, the first step of our post-training is to continue the pre-training (CPT) of a base open-weight LLM model with a standard next-token prediction objective. This process covers 27 languages/dialects\footnote{Complete list of languages available in Appendix \ref{sec:languages}.} and 47 language pairs, totaling 32B tokens. Then, before proceeding to the next phase, we merge the CPT checkpoint back with the base checkpoint. Our ablations show that merging back to the base checkpoint can improve general performance with little impact on translation quality.

\subsection{Supervised Fine-tuning}

For SFT, we collect instructions from several publicly available datasets, including OpenHermes-2.5, Aya~\cite{singh-etal-2024-aya}, Daring-Anteater~\cite{wang2024helpsteer}, Magpie~\cite{xu2024magpie}, T\"ulu~\cite{lambert2025tulu3}, and others. Using Llama 3.3 70B~\cite{llama3}, we assign two scores from 1 to 5 to each instance that represent 1) an estimate of the amount of reasoning required to answer and 2) the instruction's readability.%
\footnote{The prompt used to classify the conversations can be found in Appendix Figure \ref{fig:sft-data-curation-prompt}.} %
We filter out most data where the reasoning score or readability falls below 4.%
\footnote{We do not filter out all data with low reasoning and readability because it is also important for the model to learn how to respond to poorly formulated instructions. We only keep such prompts if they are from OpenHermes-2.5 and we discard the remaining ones coming from other datasets.} %
We then collect answers from four top-performing open-weight LLMs to create a pool of ``candidate'' answers: the original, DeepSeek V3~\cite{deepseekai2025}, Qwen 2.5 72B~\cite{qwen25}, T\"ulu 3~\cite{lambert2025tulu3}, and Llama 3.3 answers. The answer we ultimately use for training is the one that ranks the highest when evaluated using a state-of-the-art general-purpose reward model, Skywork-Gemma-2-27B~\cite{liu2024skywork}. This process follows the increasingly common paradigm of distillation from multiple teacher models, where several strong open-weight LLMs provide candidate completions, and the best one is selected based on a learned reward function. Given the multilingual nature of our corpus, this approach is closely aligned with the multilingual arbitrage method proposed by~\citet{multilingualarbitrage}, which applies the same principle of multi-teacher selection to multilingual prompts.

Additionally, similar to \Towervtwo{}~\citep{rei-etal-2024-tower}, we collect data from pre-translation, translation, and post-translation tasks. Pre-translation tasks involve preprocessing steps typically performed before translation, such as grammar error correction, named entity recognition, and the removal of PII content. Translation tasks cover a broad spectrum, including sentence-level translation, style adaptation (e.g., formal vs. informal), document-level translation, and multilingual translation (single source, targets in multiple languages). Some of these data are proprietary, but the majority comes from WMT shared tasks and Flores test sets (excluding WMT 2024). Post-translation tasks focus on processes that follow translation, such as automatic post-editing and machine translation quality evaluation. As with translation tasks, part of the data comes from proprietary sources, while the rest comes from WMT shared tasks ranging from 2017 to 2023. 

The final corpus consists of 1.3 million samples, with translation tasks accounting for approximately 22\% of the total. In Figure~\ref{fig:sft-process}, we illustrate the full SFT data curation process, including the filtering steps, along with the proportion each category contributes to the final corpus.

\begin{figure*}[h!]
    \centering
    \includegraphics[width=1.0\linewidth,trim={7cm 5.5cm 7cm 5.5cm},clip]{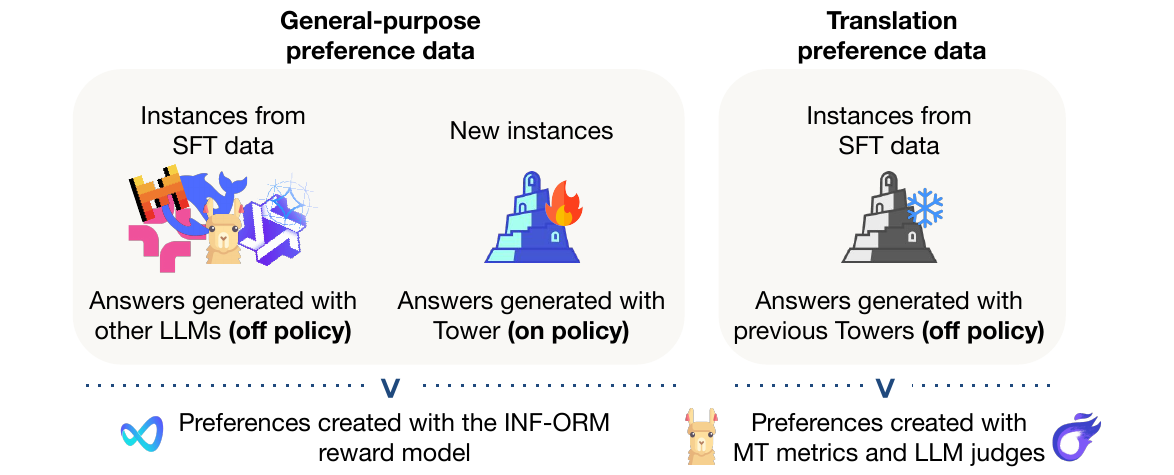}
    \caption{Process for creating and curating data for our final dataset for PO.}
    \label{fig:rl-process}
\end{figure*}

\subsection{Preference Optimization}

After SFT, our models undergo offline reinforcement learning using weighted preference optimization (WPO)~\cite{zhou-etal-2024-wpo}. This phase uses a mixture of prompts from two sources: (i) a subset of SFT prompts, and (ii) new prompts from UltraFeedback~\cite{tunstall2024zephyr}. These two datasets serve complementary roles. The SFT-derived prompts are richer in multilingual coverage, safety-critical scenarios, and multiturn interactions---areas underrepresented in UltraFeedback. Preferences for these prompts are collected off-policy from several high-quality open-weight LLMs that permit commercial use.\footnote{Models used for off-policy data: \textsc{DeepSeek V3}, \llamathree{}-70B, \qwenseventy{}-72B, \textsc{Mistral-Small-3.1}, \gemmatwoxxl{}-27B, \textsc{T"ulu-3-70B}.}

UltraFeedback data, in contrast, is used on-policy, leveraging samples generated directly by our model. For both sources, we apply the INF-ORM reward model\footnote{\url{https://huggingface.co/infly/INF-ORM-Llama3.1-70B}} to identify the best and worst completions, which are then used for WPO updates. While we experimented with alternative reward models, such as Skywork (used in the SFT phase), we observed that Gemma 2–based reward models tended to over-prefer completions from their own base models (e.g., Gemma 2 27B Instruct). This bias made them less suitable for preference optimization, where candidate responses include outputs from Gemma models. In contrast, INF-ORM, which ranked first on RewardBench at the time of writing, showed no such preference for its own Llama 3 base and was therefore selected for this phase.

Using Skywork during SFT, however, did not pose the same issue, as none of the candidate completions used to build the final SFT dataset came from Gemma 2 models. Additionally, given the large scale of the SFT data and the significantly higher inference cost of INF-ORM, we opted for Skywork.

Finally, to improve performance on machine translation, we incorporate human preference data collected by professional linguists. This data comes from two sources: (i) post-edits of \textsc{Tower v2} outputs, where the edited version is treated as preferred over the original system output, and (ii) direct preference annotations between translation variants produced during previous quality evaluations of earlier \textsc{Tower} models. We experimented with various formats and reused the collected data. Using post-edits as preference data has been shown to effectively improve translation quality~\citep{berger-etal-2024-post}. This approach is similar in spirit to the methodology described in the \textsc{Llama 3} technical report~\citep{llama3}, where expert-edited outputs were repurposed as preference data (albeit for non-MT tasks).

The remaining MT preferences are collected using \textsc{Comet22}~\cite{rei-etal-2022-comet} for MBR\footnote{We sample 24 candidates using temperature of $1.0$.} and picking the `best' and `worse' translations from the resulting MBR scores. To avoid metric-specific biases~\cite{kocmi-etal-2024-findings, pombal2025mint} we then double check that `best'  > `worse' using \metricx{}~\cite{juraska-etal-2024-metricx} and Llama 3.3 focusing on fluency and instruction following.\footnote{While neural MT metrics capture adequacy we found that an LLM-as-a-judge can better score fluency and how well the translation respects the provided user instruction.}

Regarding the choice of the optimization algorithm we found WPO to outperform DPO~\cite{rafailov_dpo} specially in terms of translation quality where we saw little to no improvement over the SFT model.

The complete reinforcement learning pipeline is summarized in Figure~\ref{fig:rl-process}.

\subsection{RL with Verifiable Rewards}
At this stage, our models already demonstrate state-of-the-art performance in both translation and general-purpose tasks. However, we find that their instruction-following, mathematical, and reasoning capabilities can be further improved through training on data with verifiable rewards. To this end, we leverage the T\"ulu 3 verifiable rewards dataset~\cite{lambert2025tulu3} and augment it with two translation-specific signals: \textit{translation-verifiable instruction} and \textit{translation preference evaluation}. We provide a template in Appendix~\ref{app:verifiable_translation_instructions}.

Translation-verifiable instructions target the model's ability to apply text transformations during translation (e.g., converting date formats from DD-MM-YYYY into MM-DD-YYYY formatting). To generate this training data, we first defined a list of 28 broad text transformations~(e.g., email formatting, date formatting, etc.; we provide all categories in Appendix~\ref{app:verifiable_translation_instructions}). Along with each transformation category, we include corresponding transformations~(e.g., for date formatting, some transformations include month abbreviation, day of week abbreviation, timezone format, etc.). These transformations come with a description, a verification (in the form of a regular expression), and one example of input/output. We show one such transformation template in Appendix~\ref{app:verifiable_translation_instructions}.

\begin{figure*}[t]
\centering
\begin{tcolorbox}[
  colback=gray!10,
  colframe=black,
  boxrule=0.5mm,
  arc=2mm,
  title=\bfseries Example of a verifiable translation instruction,
  fonttitle=\bfseries,
  width=\textwidth,
  left=2pt, right=2pt, top=4pt, bottom=4pt,
  listing only,
  left=2pt, right=2pt, top=4pt, bottom=4pt,
  listing options={style=JSON}
]
\textbf{Metadata for Prompt}\\
Length: 1 sentence\\
Topic: Economic Policy - Tax Reform\\
Guideline Category: Date Formatting: DATE\_001$^\dagger$\\

\textbf{Source Text}

The new tax reform bill, announced on January 10th, 2024, is expected to have a significant impact on the economy, with major corporations already adjusting their financial plans in anticipation of the changes that will take effect on February 20, 2025.

\textbf{Guideline}

Convert dates to MM/DD/YYYY.
\end{tcolorbox}
\caption{Examples of a verifiable translation instruction. $^\dagger$The guideline category "DATE\_001" is shown in Appendix~\ref{app:verifiable_translation_instructions}.}
\label{fig:translation-generation-template}
\end{figure*}

We then prompted  LLaMA 3.3 70B Instruct~\citep{llama3} to generate a list of precise source-dependent transformation guidelines and a source document (that does not follow them) given (i) a list of guideline categories (whose length was sampled uniformly between 1 and 4), description and one example of input/output transformation when required~(we show examples of such information in Appendix), (ii) a desired length (1/2 sentences, 1 paragraph), and (iii) a topic/sub-topic pair out of a list of over 625 pairs~(topics vary wildly spanning from pairs like "Sports Industry---Athletic Equipment" to "Journalists and Writers---Ezra Klein"). We provide the prompt for this step in Appendix~\ref{app:verifiable_translation_instructions} and one example in Figure~\ref{fig:translation-generation-template}. Next, we used the same LLM to verify the generated data. A sample was kept only if its source text violated all of the associated guidelines, ensuring every transformation was applicable. Finally, we translated these sentences using \textsc{Tower} and asked different LLMs to apply the transformations on the translated output. We filtered out any examples where the verification (e.g., regex template) did not match or where the final translation quality (measured by \textsc{CometKiwi}) was below 0.8. During GRPO, the model is rewarded when its output matches the regex in the target translation. This task is designed to encourage more precise instruction-following during translation.

Translation preference evaluation reuses the curated translation preferences from the WPO phase. Here, we prompt the model to compare two translation outputs, provide a quality assessment (reasoning), and deliver a final judgment. The model is rewarded when it selects the better translation. Note that this is done without any thinking tokens. All the reasoning that leads to the final decision is part of the final answer.

\begin{table*}[ht]
\begin{center}
\footnotesize
\begin{tabular}{lcccccccc}
    \toprule
    \multirow{2}{*}{\textbf{Models}} & \multirow{2}{*}{\textbf{Params}} & \textbf{M-ArenaHard} & \textbf{IFEval} & \multicolumn{3}{c}{\textbf{\wmttwofour}} & \multicolumn{2}{c}{\textbf{IF-MT}} \\
        \cmidrule(lr){5-7} \cmidrule(lr){8-9}
         &  &  &  & 7 lang. & 15 lang. & 24 lang. & IF & MT \\
    \midrule
    \multicolumn{7}{l}{\small \bf Closed}\\
    \gptfouro & >100B & 61.19 & 85.20 & \textbf{86.69} & \textbf{84.33} & \textbf{85.21} & \textbf{5.81} & \textbf{89.35}  \\
    \claudethreeseven & >100B & \textbf{67.00} & \textbf{89.95} & 86.41 & \textbf{84.24} & \textbf{85.19} & --- & --- \\
    \cdashlinelr{1-9}
    \multicolumn{7}{l}{\small \bf Open Weights} \\
    \almar $\dagger$ & 13B & 0.2 & 0.0 & 80.97 & -- & -- & 1.71 & 78.11 \\
    \gemmax $\dagger$ & 9B & 0.02 & 0.17 & 84.26 & 78.74 & 75.66 & 1.52 & 68.95 \\
    \Towervtwo $\dagger$ & 70B & 4.01 & 51.22 & 86.40 & \textbf{83.88} & \textbf{83.74} & 3.14 & 87.82 \\
    \gemmatwoxl & 9B & 13.38 & 66.86 & 81.93 & 75.35 & 76.34 & 5.07 & 88.51 \\
    \gemmatwoxxl & 27B & 22.81 & 66.60 & 83.68 & 79.02 & 80.18 & 5.29 & 88.67 \\
    \qwenseventy & 72B & 50.00 & 88.44 & 84.72 & 77.44 & 76.62 & 5.49 & 88.79 \\
    \llamathree & 70B & 13.15 & \textbf{92.17} & 82.74 & 78.30 & 79.48 & 5.38 & 88.13 \\
    \cdashlinelr{1-9}
    \multicolumn{7}{l}{\small \bf Ours} \\
    \textsc{Tower+} & 2B & 6.33 & 67.32 & 81.88 & 78.42 & 79.13 & 2.90 & 87.65 \\
    \textsc{Tower+} & 9B & 33.47 & 83.84 & 86.25 & 83.57 & 84.38 & 4.85 & 88.51 \\
    \textsc{Tower+} & 72B & \textbf{54.52} & 89.02 & \textbf{86.68} & 83.29 & \textbf{83.74} & \textbf{5.55} & \textbf{88.95} \\
    \bottomrule
\end{tabular}

\end{center}
\caption{Results of several translation-specific ($\dagger$) and general-purpose open-weight and closed API models across M-ArenaHard, IFEval, \wmttwofour{}, and IF-MT (English$\rightarrow$Chinese). We consider two evaluation dimensions on IF-MT: instruction-following (IF) and raw MT quality (MT). For \wmttwofour{} we report \xcomet{} and we split the language pairs into three categories: (1) seven high-resource languages, (2) the 15 languages from \Towervtwo~(our submission to WMT24), and (3) all languages supported by our new models. This categorization enables a more equitable comparison with other systems, which, in all cases, support at least the seven high-resource languages. We boldface the best overall system, and the best open-weight system if the former is proprietary.}
\label{tab:main_results}
\end{table*}

\section{Experimental Setup}

\subsection{Evaluation Setup}

Our primary objective is to achieve state-of-the-art machine translation (MT) performance and improve on general capabilities over the base model. To evaluate this, we use three benchmarks that span these two dimensions: \wmttwofour{}~\cite{deutsch2025wmt24expandinglanguagecoverage} for translation, and M-ArenaHard~\cite{dang2024ayaexpanse, li2024arena} and IFEval~\cite{zhou2023IFEval} for general-purpose performance.

\paragraph{WMT24++}
For translation evaluation, we use the WMT24++ test set, which extends the official WMT24 set~\citep{kocmi2024findings} to cover 55 languages and dialects by collecting new human-written references. The WMT shared task is a major annual competition in the field.\footnote{With over fifteen editions, WMT has become one of the flagship events at *ACL conferences.} Each year, organizers curate new data to build a test set spanning diverse domains and languages.

Since WMT24++ includes all 22 source languages covered by our models, we evaluate translation into 24 target variants.\footnote{We include both Brazilian and European Portuguese, as well as Simplified and Traditional Chinese, totaling 24 language directions. While our models support more variants, WMT24++ does not currently provide references for all of them.} 

For evaluation, we rely on state-of-the-art MT-specific automatic metrics, including \xcomet{}~\citep{guerreiro-etal-2024-xcomet} as our primary metric, and \metricx{}~\cite{juraska-etal-2024-metricx} and \chrf{}~\cite{popovic-2015-chrf} for supplementary analysis. The latter two are reported in the appendix (Tables \ref{tab:metricx_results} and \ref{tab:chrf_results} respectively).

\paragraph{IFEval} Many real-world applications of LLMs require following instructions to complete specific tasks (e.g., formatting text according to given guidelines). Instruction-following is therefore a key component of evaluating general-purpose capabilities.

To assess this, we use IFEval~\cite{zhou2023IFEval}, a widely adopted benchmark for evaluating instruction-following behavior. IFEval consists of 541 instructions whose outputs can be automatically verified using simple code or regular expressions. Models are evaluated based on the percentage of instructions executed correctly.

\paragraph{M-ArenaHard} Traditional benchmarks for evaluating general capabilities, such as MMLU~\citep{hendrycks2020measuring}, typically rely on multiple-choice questions, which limit diversity and fail to capture real-world complexity.

To address these limitations, the Chatbot Arena~\citep{zheng2023judging} adopts a more realistic evaluation setup by crowdsourcing open-ended prompts and collecting pairwise human preferences over model responses. Its main advantage is its alignment with real-world usage, where users issue diverse, complex queries. With over one million interactions, it has become the \textit{de facto} standard for evaluating general-purpose LLM performance. However, its public and slow evaluation process makes it unsuitable for rapid model development.

To mitigate this, \citet{li2024arena} proposed ArenaHard curated set of 500 representative prompts from Chatbot Arena—where model performance is measured via win rates against a fixed baseline, using an LLM-based judge rather than human annotators. They show that rankings on ArenaHard strongly correlate with those from Chatbot Arena.

In this work, we use M-ArenaHard~\citep{dang2024ayaexpanse}, a multilingual extension of ArenaHard, to evaluate general capabilities across five languages: English, German, Spanish, Chinese, and Russian. We employ LLaMA 3.3 70B Instruct~\citep{llama3} as the evaluator and use Qwen2.5 72B~\citep{qwen25} Instruct as the baseline reference model.\footnote{Using Qwen2.5 72B Instruct as baseline also gives us a direct comparison to the Instruct version of Qwen2.5 72B which we study in this paper as base model to \Towerpluslarge{}}

\paragraph{IF-MT: translation + instruction-following}
Many real-world translation tasks go beyond simple language conversion, often requiring adherence to specific guidelines and rules—such as maintaining consistent terminology, adapting date and currency formats, or converting units of measurement to align with local standards.
Thus, it is important for translation models to be capable of translating text, while having to follow instructions, which requires a mix of translation and general capabilities.
However, no existing benchmarks evaluate both dimensions, so we create one following the zero-shot benchmarking methodology~\citep[ZSB]{pombal2025zero}: IF-MT.

ZSB is a task-agnostic framework for automatically creating benchmarks that correlate strongly with human rankings by prompting language models for both data generation and evaluation.
It requires the creation of a \textit{meta prompt} for generating test instances (containing a description and some attributes of the task), and a \textit{judgement prompt}, for evaluation.

We generate test data for two LPs---English$\rightarrow$Chinese and English$\rightarrow$Spanish (Latin America)---using \claudethreeseven{}.
We do not consider a fixed set of instructions, but rather prompt the data generator model to come up with 2 to 4 instructions that can be applied to the source text it generates.\footnote{We consider only ``verifiable'' instructions---e.g., currency/date formatting, glossary following---as opposed to subjective ones, like style guides.}

For evaluation, we disentangle translation quality from instruction-following by considering two metrics: 1) \comet{}-22~\citep{rei-etal-2022-comet}, a state-of-the-art MT metric with a larger context length than \xcomet{} (our generated sources are large enough that this is an issue); and 2) \claudethreeseven{}{} as a judge for evaluating the extent to which models follow the instructions.
On the latter, the judge scores each instance from 1 (worst) to 6 (best), as specified in the judgment prompt.
LLMs have been shown to be good evaluators of these capabilities~\citep{zheng2023judging,zengevaluating,pombal2025zero}.

For each model and evaluation dimension, we report the average score over all instances.
We omit \claudethreeseven{} from the results since its performance would be overestimated due to intra-model family bias.
In Appendix~\ref{sec:zsb}, we include the data generation and judgment prompts we used, two examples from our benchmark (one for each LP), and results for English$\rightarrow$Chinese (conclusions are similar across LPs).

\paragraph{Baselines} 
We evaluate our models against both closed-source API models and open-weight LLMs. For closed models, we include \gptfouro{} and \claudethreeseven{}. Among open-weight models, we consider leading general-purpose LLMs with fewer than 80B parameters, as well as translation-focused models such as \almar{}~\cite{xualmar}, \gemmax{}~\cite{gemmax}, and \Towervtwo{}~\citep{rei-etal-2024-tower}, the winning submission of the WMT24 MT shared task. For most models, we use a standardized prompting format (see Fig. \ref{fig:translation-prompt} in Appendix); exceptions include translation-specific models, where we adopt the prompts recommended by their respective authors.

\subsection{Main Results}
From Table~\ref{tab:main_results}, we observe that the new \textsc{Tower} models achieve a strong balance between translation performance and general chat capabilities. \Towerpluslarge{} achieves competitive results on instruction-following benchmarks (IFEval), performing on par with leading models such as \claudethreeseven{} and \gptfouro{}, and surpassing strong open-weight baselines like \qwenseventy{} on M-ArenaHard. Notably, \Towerpluslarge{} matches the translation performance of the previous \Towervtwo{} model while substantially improving win rates against \qwenseventy{} on M-ArenaHard, from 4\% to 54.5\%.
On IF-MT, \Towerpluslarge{} once again surpasses all other open models on both evaluation dimensions, showcasing its ability to leverage both translation and general capabilities.

Meanwhile, \Towerplusmedium{}, despite having only 9B parameters, achieves competitive performance across 24 language pairs (LPs) in machine translation and outperforms \gemmatwoxxl{} on IFEval, M-ArenaHard, and IF-MT. Finally, \Towerplussmall{}, our smallest model with just 2B parameters, matches the machine translation performance of \llamathree{} and outperforms the previous \Towervtwo{}-70B model on both M-ArenaHard, IFEval, and the instruction-following side of IF-MT, highlighting the effectiveness of our improved post-training pipeline.

We note that, when comparing the previous \Towervtwo{}-70B model to our new 72B model (\Towerpluslarge{}), there is a slight decrease in translation quality on the subset of 15 languages originally used in WMT24. We attribute this drop not to the expanded language coverage in the new version (from 15 to 22 languages), but rather to the more limited multilingual capabilities of the \textsc{Qwen 2.5} base model. This limitation is also evident when comparing \textsc{Qwen 2.5} 72B Instruct with \textsc{Llama 3.3} 70B: while \textsc{Qwen 2.5} 72B Instruct performs strongly on general chat capabilities (with a win rate of 86.85\% over \textsc{Llama 3.3}) and excels in translation for high-resource languages, its performance sharply declines when evaluated across a broader set of 15 or 22 languages.

We further analyze the impact of base model selection in Section~\ref{sec:base_models}. Although \textsc{Llama 3} models exhibit stronger translation capabilities, their more restrictive licensing---including mandatory attribution and naming requirements---led us to prioritize \textsc{Qwen 2.5} and \textsc{Gemma 2} for this work.

All MT-specific models perform poorly on IF-MT.
In fact, we had to remove the instructions from the prompt of \almar{} and \gemmax{} so that the models were able to translate at all, highlighting the need to create more flexible models specialized on MT.\footnote{We also noticed that the performance of these two models was hurt because they often failed to translate all paragraphs in sources with line breaks (e.g., they stopped generating tokens after the first or second paragraph). This inability to translate long sources with line breaks---due to overfitting to a specific format---can be seen as another shortcoming of inflexible MT-specific models.}
Crucially, \Towerpluslarge{} greatly outperforms \Towervtwo{} on this benchmark, which further speaks to the effectiveness of our approach in balancing translation quality and general capabilities.

\subsection{Impact of Different Training Stages}
\label{sec:impact_stages}

\begin{figure}[t]
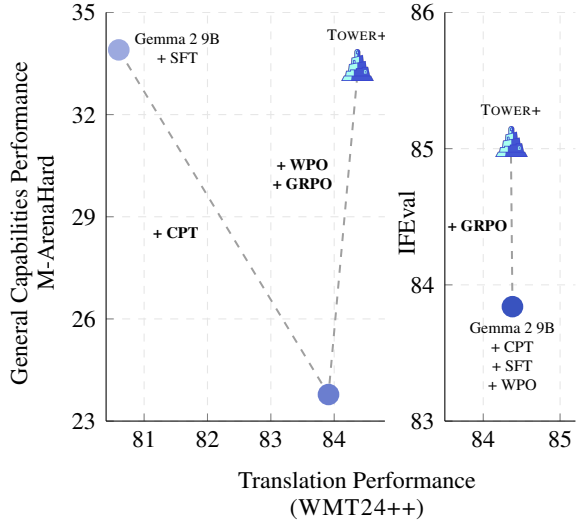

    \begin{subfigure}[t]{0.65\columnwidth}
    \vspace{0pt}
    \centering
    \pgfplotsset{
        footnotesize,
        samples=10,
        xmin=80.4, xmax=84.8,
        ymin=23, ymax=35,
        xtick={80,81,82,83,84,85},
        xticklabels={80,81,82,83,84,85},
        ytick={23,26,29,32,35},
        yticklabels={23,26,29,32,35},
        grid style={dashed,color=gray!20},
        grid=both,
        xlabel=Translation Performance \\ (\wmttwofour{}),
        x label style={at={(axis description cs:0.90,-0.1)},anchor=north},
        ylabel=General Capabilities Performance \\ M-ArenaHard,
        y label style={at={(axis description cs:-0.15,0.485)},anchor=south},
    }
    
    \begin{tikzpicture}
    \begin{groupplot}[
        group style = {group size = 1 by 1, horizontal sep = 24pt},
        width = 5.25cm,
        height = 7cm
    ]
        \nextgroupplot[
            align = center,
            legend style={
                at={(0.5,1.25)},
                anchor=north,
                legend columns=2,
                column sep=10pt,
                draw=none,
            },
            legend cell align={left},
            axis x line*=bottom,
            axis y line*=left,
            y label style={at={(axis description cs:-0.15,0.485)},anchor=south},
            xtick pos=bottom,
            ytick pos=left,
        ]
        \addplot[gray!75, line width=0.75pt, dashed, forget plot] coordinates {(80.60, 33.9) (83.91, 23.78) (84.38, 33.47) (84.36, 33.3)};

        \addplot[color=CustomBlue!50, line width=1.1pt, mark=*, mark size=3.5pt] coordinates {(80.60, 33.9)};
        \node[anchor=north, yshift=10pt, xshift=22pt] at (axis cs:80.60, 33.9) {\tiny\shortstack{Gemma 2 9B\\+ SFT}};

        \node[anchor=north] at (axis cs:81.5, 29) {\tiny\shortstack{\textbf{+ CPT}}};

        \addplot[color=CustomBlue!75, line width=1.1pt, mark=*, mark size=3.5pt] coordinates {(83.91, 23.78)};

        \node[anchor=north] at (axis cs:83.5, 31) {\tiny\shortstack{\textbf{+ WPO} \\ \textbf{+ GRPO}}};

        \node at (axis cs:84.36, 33.3) [anchor=south west, yshift=-7.5pt, xshift=-9.5pt] {\includegraphics[height=0.4cm]{latex/logos/tower.png}};
        \node[anchor=north] at (axis cs:84.36, 34.8) {\tiny{\textsc{Tower+}}};


    \end{groupplot}
    \end{tikzpicture}
    \end{subfigure}
    \begin{subfigure}[t]{0.3\columnwidth}
    \vspace{0pt}
    \centering
    \pgfplotsset{
        footnotesize,
        samples=10,
        xmin=83.5, xmax=85.2,
        ymin=83, ymax=86,
        xtick={84,85},
        xticklabels={84,85},
        ytick={83,84,85,86},
        yticklabels={83,84,85,86},
        grid style={dashed,color=gray!20},
        grid=both,
        xlabel=T \\ W,
        x label style={at={(axis description cs:0.5,-0.1)},anchor=north,color=white},
        ylabel=IFEval,
        y label style={at={(axis description cs:-0.5,0.485)},anchor=south},
    }
    
    \begin{tikzpicture}
    \begin{groupplot}[
        group style = {group size = 1 by 1, horizontal sep = 24pt},
        width = 3.3cm, 
        height = 7cm
    ]
        \nextgroupplot[
            align = center,
            legend style={
                at={(0.5,1.25)},
                anchor=north,
                legend columns=2,
                column sep=10pt,
                draw=none,
            },
            legend cell align={left},
            axis x line*=bottom,
            axis y line*=left,
            y label style={at={(axis description cs:-0.15,0.485)},anchor=south},
            xtick pos=bottom,
            ytick pos=left,
        ]
        \addplot[gray!75, line width=0.75pt, dashed, forget plot] coordinates {(84.38, 83.84) (84.36, 85.01)};

        \addplot[color=CustomBlue, line width=1.1pt, mark=*, mark size=3.5pt] coordinates {(84.38, 83.84)};
        \node[anchor=north] at (axis cs:84.38, 83.8) {\tiny\shortstack{Gemma 2 9B\\+ CPT\\+ SFT\\+ WPO}};

        \node[anchor=north] at (axis cs:83.925, 84.55) {\tiny{\textbf{+ GRPO}}};

        \node at (axis cs:84.36, 85.01) [anchor=south west, yshift=-7.5pt, xshift=-9.5pt] {\includegraphics[height=0.4cm]{latex/logos/tower.png}};
        \node[anchor=north] at (axis cs:84.36, 85.4) {\tiny{\textsc{Tower+}}};

    \end{groupplot}
    \end{tikzpicture}
    \end{subfigure}
\caption{Performance comparison across progressive training stages using \textsc{Gemma 2 9B} as the foundation model. We represent a total of 4 training setups: (1) SFT (base): Supervised Fine-Tuning directly on the base model; (2) CPT+SFT: Continued Pre-Training followed by SFT; (3) CPT+SFT+WPO: Addition of Preference Optimization; and (4) CPT+SFT+WPO+GRPO: Integration of GRPO with verifiable rewards. The left plot omits (3) because performance is identical to (4) on M-ArenaHard; GRPO brings gains on IFEval. 
}
    \label{fig:impact_diff_stages}
\end{figure}

Our first ablation study examines the impact of each stage in our post-training pipeline on achieving a strong balance between general capabilities and translation quality. To this end, we perform the following comparisons:
\begin{enumerate}
    \item \textbf{Impact of CPT:} We take a \textsc{Gemma 2 9B} model and run only the SFT stage without any CPT. This allows us to isolate and measure the effect of the CPT phase on both general capabilities and translation performance.
    \item \textbf{Impact of WPO:} After establishing the contribution of CPT, we evaluate the gains introduced by the WPO stage.
    \item \textbf{Impact of GRPO:} Finally, we assess the effect of Reinforcement Learning with Verifiable Rewards by comparing the SFT+WPO model to our full pipeline (CPT+SFT+WPO+GRPO).
\end{enumerate}

Figure~\ref{fig:impact_diff_stages} summarizes model performance at each training stage, using both general-purpose benchmarks and translation-specific evaluations based on \xcomet{}.

The CPT phase primarily improves translation quality, particularly for mid- and low-resource languages. While it increases \xcomet{} by only 0.77 points on the 7 high-resource language pairs, it delivers an overall gain of 3.3 points when evaluating across all languages. However, this improvement comes at a cost: we observe a consistent degradation in general capabilities as measured by M-ArenaHard. Although the exact cause is difficult to pinpoint without full access to the base model's training details, we hypothesize that this degradation stems from disrupting the delicate balance achieved during the final pretraining annealing phases. These phases often involve carefully curated data, gradual learning rate schedules, and internal optimizations that are difficult to reproduce. Restarting training—even with high-quality data—may shift the data distribution away from key domains~\cite{wang2025learningdynamicscontinualpretraining} such as math, code, or STEM, or reset optimizer states in a way that hurts general capabilities. Nonetheless, given our primary goal of maximizing translation quality, we consider the observed gains in multilingual performance to outweigh the loss on M-ArenaHard.\footnote{As open-weight models become increasingly multilingual, and given that CPT requires significantly more compute than subsequent stages, we question the necessity of CPT in future pipelines.} 

The WPO stage contributes substantial improvements across instruction following, general chat ability, and translation, confirming its central role in aligning the model.

While GRPO appears promising, it is the stage where we observed the least overall gains. Improvements were primarily limited to IFEval, which aligns with the fact that the T\"ulu 3 dataset is specifically designed to target IFEval and GSM8k. However, more indirect signals—such as translation performance—showed no measurable improvement.

Although the gains on IFEval are initially encouraging, our analysis suggests they may stem from overfitting to the benchmark. In particular, we found that several prompts in the T\"ulu 3 dataset are poorly formatted, yet the verifiable reward still expects exact compliance with the instructions (see Figure~\ref{fig:tulu-3-bad-prompt}). After cleaning the dataset to remove such inconsistencies, GRPO no longer yielded improvements over the WPO-only model.

These findings suggest that while GRPO with verifiable rewards remains a promising strategy for improving targeted model capabilities, its broader effectiveness depends heavily on the quality and structure of the reward-aligned data. Further research is needed to better integrate GRPO with VR into our post-training pipeline in a way that generalizes beyond narrow benchmarks like IFEval.

\subsection{Importance of base model}
\label{sec:base_models}

\begin{figure}[t]
    \centering
    \pgfplotsset{
        footnotesize,
        samples=10,
        xmin=57, xmax=87,
        ymin=10, ymax=40,
        xtick={60,65,70,75,80,85},
        xticklabels={60,65,70,75,80,85},
        ytick={10,15,20,25,30,35,40},
        yticklabels={10,15,20,25,30,35,40},
        grid style={dashed,color=gray!20},
        grid=both,
        xlabel=Translation Performance (\wmttwofour{}),
        x label style={at={(axis description cs:0.5,-0.1)},anchor=north},
        ylabel=General Capabilities Performance \\ (M-ArenaHard),
        y label style={at={(axis description cs:-0.15,0.485)},anchor=south},
    }
    
    \begin{tikzpicture}
    \begin{groupplot}[
        group style = {group size = 1 by 1, horizontal sep = 24pt},
        width = \columnwidth, 
        height = 7cm
    ]
        \nextgroupplot[
            align = center,
            legend style={
                at={(0.5,1.25)},
                anchor=north,
                legend columns=2,
                column sep=10pt,
                draw=none,
            },
            legend cell align={left},
            axis x line*=bottom,
            axis y line*=left,
            y label style={at={(axis description cs:-0.09,0.485)},anchor=south},
            xtick pos=bottom,
            ytick pos=left,
        ]
        \addplot[orange!50, line width=0.75pt, -{Latex[length=3pt]}, forget plot, mark=*, mark indices={1}, mark options={fill=orange!50}] coordinates {(69, 34) (61, 34)};
        \node[anchor=south, font=\tiny] at (axis cs:69, 34.25) {High-resource \\ languages};
        \node[anchor=south, font=\tiny] at (axis cs:61, 34.25) {All \\ languages};
        \addplot[orange!50, line width=0.75pt, forget plot] coordinates {(60.26, 19.87) (76.95, 19.87)};
        \node at (axis cs:76.95, 19.87) [anchor=south west, yshift=-9.5pt, xshift=-9.5pt] {\includegraphics[height=0.4cm]{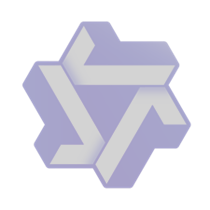}};
        \node at (axis cs:60.26, 19.87) [anchor=south west, yshift=-9.5pt, xshift=-9.5pt] {\includegraphics[height=0.4cm]{latex/logos/qwen.png}};
        \node[anchor=north, yshift=14pt] at (axis cs:60.26, 19.87) {\tiny{Qwen 2.5 7B}};

        \addplot[orange!50, line width=0.75pt, forget plot] coordinates {(63.96, 19.38) (79.10, 19.38)};
        \node at (axis cs:79.10, 19.38) [anchor=south west, yshift=-9.5pt, xshift=-9.5pt] {\includegraphics[height=0.4cm]{latex/logos/qwen_gray.png}};
        \node at (axis cs:63.96, 19.38) [anchor=south west, yshift=-9.5pt, xshift=-9.5pt] {\includegraphics[height=0.4cm]{latex/logos/qwen.png}};
        \node[anchor=north, yshift=-2pt] at (axis cs:63.96, 19.38) {\tiny{Qwen 2.5 14B}};

        \addplot[orange!50, line width=0.75pt, forget plot] coordinates {(76.34, 13.38) (81.93, 13.38)};
        \node at (axis cs:81.93, 13.38) [anchor=south west, yshift=-9.5pt, xshift=-9.5pt] {\includegraphics[height=0.4cm]{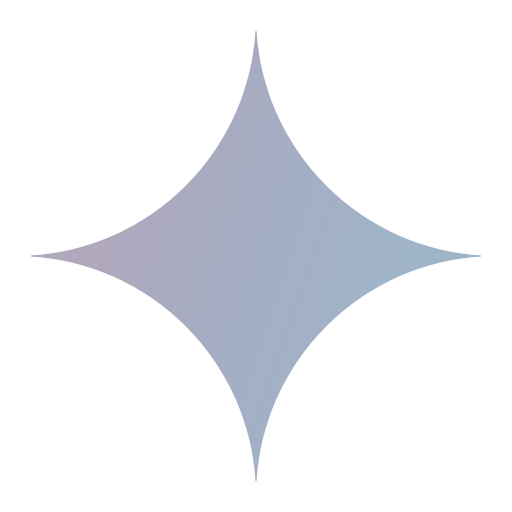}};
        \node at (axis cs:76.34, 13.38) [anchor=south west, yshift=-9.5pt, xshift=-9.5pt] {\includegraphics[height=0.4cm]{latex/logos/gemini.png}};
        \node[anchor=north, yshift=-1pt] at (axis cs:76.34, 13.38) {\tiny{Gemma 2 9B}};

        \addplot[orange!50, line width=0.75pt, forget plot] coordinates {(83.91, 23.78) (85.85, 23.78)};
        \node at (axis cs:85.85, 23.78) [anchor=south west, yshift=-7.5pt, xshift=-9.5pt] {\includegraphics[height=0.3cm]{latex/logos/tower_gray.png}};
        \node at (axis cs:83.91, 23.78) [anchor=south west, yshift=-7.5pt, xshift=-9.5pt] {\includegraphics[height=0.3cm]{latex/logos/tower.png}};
        \node[anchor=north, yshift=-1pt] at (axis cs:83.91, 23.78) {\tiny\shortstack{\textsc{Tower+}\\Gemma 2 9B}};

        \addplot[orange!50, line width=0.75pt, forget plot] coordinates {(77.48, 29.76) (83.37, 29.76)};
        \node at (axis cs:83.37, 29.76) [anchor=south west, yshift=-7.5pt, xshift=-9.5pt] {\includegraphics[height=0.3cm]{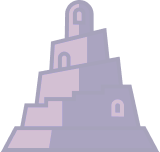}};
        \node at (axis cs:77.48, 29.76) [anchor=south west, yshift=-7.5pt, xshift=-9.5pt] {\includegraphics[height=0.3cm]{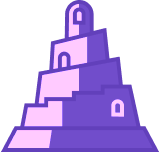}};
        \node[anchor=north, yshift=-1pt] at (axis cs:77.48, 29.76) {\tiny\shortstack{\textsc{Tower+}\\Qwen 2.5 7B}};

        \addplot[orange!50, line width=0.75pt, forget plot] coordinates {(80.12, 37.18) (84.89, 37.18)};
        \node at (axis cs:84.89, 37.18) [anchor=south west, yshift=-7.5pt, xshift=-9.5pt] {\includegraphics[height=0.3cm]{latex/logos/tower_pink_gray.png}};
        \node at (axis cs:80.12, 37.18) [anchor=south west, yshift=-7.5pt, xshift=-9.5pt] {\includegraphics[height=0.3cm]{latex/logos/tower_pink.png}};
        \node[anchor=north, yshift=-1pt] at (axis cs:80.12, 37.18) {\tiny\shortstack{\textsc{Tower+}\\Qwen 2.5 14B}};

    \end{groupplot}
    \end{tikzpicture}
\caption{Comparison of \textsc{Gemma 2 9B} and \textsc{Qwen 2.5 7B/14B} on translation quality (WMT24++) and general chat capabilities (M-ArenaHard). Arrows show the performance drop when including all languages vs high-resource ones.}
\label{fig:base_model_tradeoff}
\end{figure}

Our second ablation examines the importance of base model selection. Since our goal is to balance strong translation support with general-purpose capabilities, we focus on two prominent model families: \textsc{Qwen 2.5} and \textsc{Gemma 2}. While \textsc{Qwen 2.5} models demonstrate outstanding performance across a range of general-purpose benchmarks, they exhibit limited multilingual capabilities and comparatively weaker performance on translation tasks~\cite{gemmax}. In contrast, the \textsc{Gemma 2} family achieves state-of-the-art machine translation performance among open-weight models while maintaining competitive results on general tasks. By comparing these two model families, we directly study the trade-off between a more multilingual-oriented base model (\textsc{Gemma 2}) and a model optimized for general-purpose capabilities but less robust in multilingual settings (\textsc{Qwen 2.5}).

To better understand this trade-off, we compare \textsc{Qwen 2.5} models (both 7B and 14B) with \textsc{Gemma 2 9B} by running the first two stages of our pipeline (CPT and SFT). Figure~\ref{fig:base_model_tradeoff} illustrates the results. As shown, while \textsc{Qwen 2.5} models demonstrate stronger capabilities on M-ArenaHard, they consistently lag behind in translation quality, particularly when mid- and low-resource languages are included. Notably, even with a larger parameter count, \textsc{Qwen 2.5 14B} fails to match the translation performance of \textsc{Gemma 2 9B}. In contrast, for general chat capabilities, \textsc{Qwen 2.5 7B} surpasses \textsc{Gemma 2 9B} despite having fewer parameters. This trend is also reflected in the released Instruct models, where \textsc{Qwen 2.5 7B Instruct} outperforms \textsc{Gemma 2 9B Instruct} on M-ArenaHard but shows significantly weaker results on translation tasks.

\section{Conclusions}

In this paper, we presented a complete post-training pipeline designed to balance a task-specific use case—machine translation—with general-purpose language model capabilities. We demonstrated that it is possible to build models that achieve competitive translation performance while maintaining strong results on instruction-following and general chat benchmarks.

Through extensive ablations, we analyzed the contribution of each stage in our pipeline. Notably, we found that while Continued Pretraining (CPT) boosts translation performance—especially for mid- and low-resource languages—it can slightly degrade general capabilities, highlighting an important trade-off. The WPO stage consistently improves both translation and general-purpose performance, while GRPO shows targeted gains in instruction following but limited impact on broader capabilities.

Our final models not only outperform the Instruct-tuned versions released by the original model authors, but also surpass leading open-weight models such as \textsc{Llama 3} and \textsc{Qwen 2.5 72B} on general-purpose benchmarks like M-ArenaHard. In translation, our models achieve results on par with frontier systems such as \textsc{GPT-4} and \textsc{Claude 3.7}, while maintaining competitive general chat abilities.

Overall, our findings suggest that post-training pipelines can be tailored to business-specific needs, such as localization, without sacrificing broader general intelligence. Future work includes further exploration of RL with verifiable rewards to better align models across a wider range of tasks, and improving the handling of mid- and low-resource languages.

\section*{Limitations}

While our work demonstrates the effectiveness of carefully designed post-training pipelines in balancing translation performance and general capabilities, several limitations remain.

First, as open-weight models continue to improve their native multilingual capabilities, the marginal benefit of task-specific adaptations such as CPT may decrease. Although our experiments show that even relatively multilingual base models like \textsc{Gemma 2 9B} still benefit from CPT, further research is needed to understand how these dynamics evolve as base models become increasingly capable out-of-the-box.

Second, while our study covers models up to 72B parameters, it remains unclear whether similar post-training pipelines would provide sufficient gains for much larger frontier models (e.g., 100B+ parameters). As models scale up, the relative improvements from targeted post-training may diminish compared to the associated compute cost. Nonetheless, we argue that for many real-world business use cases—particularly those involving well-defined tasks like translation and localization—building more efficient, specialized models remains highly valuable and preferable to deploying very large, general-purpose systems.

Third, while our pipeline improves translation quality and general chat performance, it was specifically validated for multilingual translation tasks and does not explicitly optimize for other domain-specific areas such as biomedicine or legal reasoning. Nonetheless, we believe the overall recipe could serve as inspiration for designing similar post-training pipelines tailored to specialized domains.

Finally, while we evaluated across a wide range of languages and tasks, the available benchmarks are still skewed toward high- and mid-resource languages. Further improving performance in truly low-resource and code-switched scenarios is an important area for future exploration.

\section*{Acknowledgments}
Part of this work was supported by the EU’s Horizon Europe Research and Innovation Actions (UTTER, contract 101070631), by the project DECOLLAGE (ERC-2022-CoG
101088763), and by the Portuguese Recovery and Resilience Plan through project C645008882-00000055 (Center for Responsible AI). We thank EuroHPC for the HPC resources used to support this work through grant EHPC-EXT-2023E01-042 and grant EHPC-AI-2024A01-085.

\bibliography{custom}

\appendix

\section*{Appendix}
\label{sec:appendix}

\section{Covered languages}
\label{sec:languages}
\textsc{Tower} models presented in this paper cover the following 27 languages/dialects: German, Spanish, Spanish (Latin America), French, Italian, Korean, Dutch, Russian, English, Portuguese (Portugal), Portuguese (Brazilian), , Chinese (Simplified), Chinese (Traditional), Czech, Ukrainian, Hindi, Icelandic, Japanese, Polish, Swedish, Hungarian, Romanian, Danish, Norwegian (Nynorsk), Norwegian (Bokmål), Finnish.

Table \ref{tab:chrf_results}, \ref{tab:metricx_results} and \ref{tab:xcomet_results} show results for all supported languages from WMT24++ testset~\cite{deutsch2025wmt24expandinglanguagecoverage} using \textsc{chrF}~\cite{popovic-2015-chrf}, \textsc{MetricX24-xxl}~\cite{juraska-etal-2024-metricx} and \textsc{xComet-xxl} respectively.

\begin{sidewaystable*}
\begin{center}
\setlength{\tabcolsep}{3pt}
\scriptsize
\begin{tabular}{l>{\columncolor{avg7color!10}}c>{\columncolor{avg15color!5}}c>{\columncolor{avgAllcolor!10}}cccccccccccccccccccccccc}
    \toprule
Models & Avg. 7 & Avg. 15 & Avg. & \cellcolor{avg7color!10}pt\_BR & \cellcolor{avg15color!5}pt\_PT & \cellcolor{avg7color!10}zh\_CN & \cellcolor{avg15color!5}zh\_TW & \cellcolor{avg15color!5}cs\_CZ & \cellcolor{avgAllcolor!10}da\_DK & \cellcolor{avg15color!5}nl\_NL & \cellcolor{avgAllcolor!10}fi\_FI & \cellcolor{avg7color!10}fr\_FR & \cellcolor{avg7color!10}de\_DE & \cellcolor{avg15color!5}hi\_IN & \cellcolor{avgAllcolor!10}hu\_HU & \cellcolor{avg15color!5}is\_IS & \cellcolor{avg7color!10}it\_IT & \cellcolor{avg15color!5}ja\_JP & \cellcolor{avg15color!5}ko\_KR & \cellcolor{avgAllcolor!10}no\_NO & \cellcolor{avgAllcolor!10}pl\_PL & \cellcolor{avgAllcolor!10}ro\_RO & \cellcolor{avg7color!10}ru\_RU & \cellcolor{avg7color!10}es\_MX & \cellcolor{avgAllcolor!10}sv\_SE & \cellcolor{avg15color!5}uk\_UA \\
\hline
\gptfouro{} & 60.11 & 53.96 & 55.19 & 65.99 & 63.21 & 39.80 & 32.36 & 57.11 & 66.36 & 62.12 & 62.74 & 65.03 & 61.23 & 40.68 & 56.38 & 49.57 & 66.99 & 34.68 & 34.26 & 51.75 & 49.90 & 64.65 & 53.11 & 68.65 & 66.54 & 56.22 \\
\claudethreeseven{} & 60.93 & 55.29 & 56.60 & 66.56 & 63.21 & 40.66 & 34.14 & 58.38 & 68.13 & 62.56 & 64.99 & 65.78 & 61.64 & 40.71 & 56.93 & 50.33 & 68.13 & 38.10 & 39.19 & 54.76 & 52.26 & 65.68 & 54.72 & 69.04 & 67.61 & 58.26 \\
\cdashlinelr{1-27}
\almar{} & 48.97 & 39.98 & 36.41 & 52.99 & 50.44 & 27.67 & 13.93 & 46.09 & 20.69 & 48.88 & 31.14 & 54.02 & 54.17 & 7.84 & 29.52 & 42.05 & 52.12 & 21.47 & 11.78 & 21.20 & 33.92 & 46.73 & 46.53 & 55.33 & 30.06 & 38.77 \\
\gemmax & 57.83 & 47.58 & 43.11 & 63.76 & 59.84 & 39.12 & 19.42 & 53.19 & 44.21 & 60.26 & 17.01 & 62.00 & 59.57 & 39.35 & 15.01 & 13.07 & 65.15 & 33.34 & 34.50 & 46.38 & 50.02 & 29.30 & 50.12 & 65.07 & 44.20 & 27.63 \\
\gemmatwoxl & 57.37 & 49.77 & 50.85 & 63.53 & 60.29 & 35.98 & 30.14 & 50.69 & 61.20 & 58.21 & 55.77 & 62.64 & 58.10 & 37.94 & 48.18 & 32.75 & 64.15 & 32.59 & 31.20 & 48.83 & 47.53 & 59.43 & 50.47 & 66.72 & 61.32 & 51.77\\
\gemmatwoxxl & 58.67 & 51.69 & 52.85 & 64.78 & 61.75 & 38.61 & 32.24 & 53.13 & 63.95 & 60.34 & 57.13 & 63.62 & 59.21 & 39.05 & 51.23 & 38.60 & 65.46 & 34.32 & 32.95 & 51.48 & 48.86 & 61.77 & 51.49 & 67.50 & 63.37 & 54.65\\
\qwenseventy  & 57.60 & 50.04 & 50.62 & 64.17 & 61.22 & 40.18 & 32.92 & 50.95 & 59.35 & 59.00 & 51.74 & 63.43 & 56.62 & 36.35 & 44.18 & 34.52 & 64.21 & 33.67 & 33.31 & 48.19 & 47.74 & 58.25 & 47.99 & 66.59 & 60.14 & 49.57\\
\llamathree & 56.94 & 50.22 & 51.84 & 64.45 & 60.62 & 28.17 & 27.04 & 53.36 & 63.12 & 60.43 & 58.33 & 63.12 & 58.72 & 38.44 & 52.64 & 40.72 & 65.43 & 32.88 & 26.66 & 52.15 & 49.17 & 62.06 & 51.48 & 67.21 & 64.14 & 51.99 \\
\cdashlinelr{1-27}
\Towervtwo & 59.71 & 53.71 & 53.72 & 65.64 & 62.52 & 40.43 & 22.24 & 55.82 & 62.46 & 62.18 & 58.09 & 63.82 & 60.86 & 40.29 & 52.82 & 48.64 & 66.90 & 34.79 & 35.60 & 50.30 & 49.20 & 62.57 & 52.72 & 67.61 & 63.50 & 56.60\\
\Towerplussmall & 56.35 & 50.46 & 51.81 & 63.31 & 59.93 & 35.67 & 30.51 & 51.76 & 63.75 & 59.08 & 57.84 & 61.24 & 57.48 & 38.63 & 50.31 & 45.91 & 63.42 & 32.46 & 32.41 & 50.98 & 48.15 & 61.58 & 47.81 & 65.54 & 62.21 & 51.70\\
\Towerplusmedium & 59.02 & 53.37 & 54.60 & 65.06 & 62.28 & 39.67 & 32.75 & 55.44 & 65.63 & 60.96 & 61.05 & 63.19 & 59.75 & 40.25 & 53.78 & 48.94 & 65.56 & 36.08 & 36.34 & 53.81 & 50.23 & 63.90 & 52.31 & 67.63 & 65.02 & 56.06\\
\Towerpluslarge & 59.83 & 53.61 & 54.46 & 65.50 & 62.41 & 42.52 & 32.92 & 55.10 & 65.25 & 61.04 & 59.68 & 63.64 & 60.04 & 39.66 & 52.41 & 47.25 & 66.10 & 35.95 & 37.12 & 53.04 & 49.61 & 62.55 & 53.16 & 67.83 & 64.24 & 55.62\\
\bottomrule
\end{tabular}

\end{center}
\caption{ChrF ($\uparrow$) results for WMT24++ on all tested language pairs, including the averages across 
high-resource (in \textcolor{avg7color!30}{blue}), the 
additional languages/dialects supported by \Towervtwo{}~\cite{rei-etal-2024-tower} (in \textcolor{avg15color!30}{green}), and 
all languages supported by our new models (in \textcolor{avgAllcolor!30}{orange}). Note that these averages are cumulative meaning that, ``Avg. 15'' includes also the languages from ``Avg. 7''}
\label{tab:chrf_results}
\end{sidewaystable*}

\begin{sidewaystable*}
\begin{center}
\setlength{\tabcolsep}{3pt}
\scriptsize
\begin{tabular}{l>{\columncolor{avg7color!10}}c>{\columncolor{avg15color!5}}c>{\columncolor{avgAllcolor!10}}cccccccccccccccccccccccc}
    \toprule
Models & Avg. 7 & Avg. 15 & Avg. & \cellcolor{avg7color!10}pt\_BR & \cellcolor{avg15color!5}pt\_PT & \cellcolor{avg7color!10}zh\_CN & \cellcolor{avg15color!5}zh\_TW & \cellcolor{avg15color!5}cs\_CZ & \cellcolor{avgAllcolor!10}da\_DK & \cellcolor{avg15color!5}nl\_NL & \cellcolor{avgAllcolor!10}fi\_FI & \cellcolor{avg7color!10}fr\_FR & \cellcolor{avg7color!10}de\_DE & \cellcolor{avg15color!5}hi\_IN & \cellcolor{avgAllcolor!10}hu\_HU & \cellcolor{avg15color!5}is\_IS & \cellcolor{avg7color!10}it\_IT & \cellcolor{avg15color!5}ja\_JP & \cellcolor{avg15color!5}ko\_KR & \cellcolor{avgAllcolor!10}no\_NO & \cellcolor{avgAllcolor!10}pl\_PL & \cellcolor{avgAllcolor!10}ro\_RO & \cellcolor{avg7color!10}ru\_RU & \cellcolor{avg7color!10}es\_MX & \cellcolor{avgAllcolor!10}sv\_SE & \cellcolor{avg15color!5}uk\_UA \\
\hline
\gptfouro{} & -4.00 & -4.53 & -4.81 & -4.65 & -5.13 & -3.10 & -2.85 & -6.12 & -4.68 & -3.77 & -6.02 & -4.51 & -2.64 & -4.28 & -6.58 & -7.46 & -4.29 & -4.37 & -4.42 & -5.66 & -6.37 & -5.62 & -4.60 & -4.25 & -4.34 & -5.00 \\
\claudethreeseven{}  & -4.10 & -4.63 & -4.92 & -4.86 & -5.20 & -3.18 & -2.85 & -6.28 & -4.70 & -3.82 & -6.06 & -4.60 & -2.74 & -4.38 & -6.74 & -7.51 & -4.38 & -4.55 & -4.44 & -5.82 & -6.57 & -5.84 & -4.53 & -4.40 & -4.50 & -5.12 \\ 
\cdashlinelr{1-27}
\almar & -4.58 & -5.83 & -6.82 & -5.80 & -5.99 & -3.33 & -3.03 & -6.62 & -6.51 & -5.34 & -13.58 & -5.23 & -2.87 & -7.49 & -13.66 & -7.42 & -5.24 & -6.29 & -9.74 & -6.02 & -9.84 & -8.97 & -4.91 & -4.72 & -7.72 & -6.58 \\
\gemmax  & -4.46 & -5.10 & -5.83 & -5.33 & -5.36 & -3.54 & -2.93 & -7.10 & -6.84 & -3.98 & -9.89 & -4.83 & -3.21 & -5.21 & -10.12 & -8.16 & -4.45 & -5.36 & -5.12 & -6.31 & -7.19 & -7.48 & -5.54 & -4.32 & -6.57 & -5.25 \\
\gemmatwoxl & -4.69 & -6.05 & -6.26 & -5.33 & -5.75 & -3.60 & -3.30 & -8.00 & -5.79 & -4.74 & -8.52 & -5.22 & -3.42 & -5.33 & -8.88 & -16.47 & -4.93 & -5.26 & -5.93 & -7.00 & -7.63 & -6.90 & -5.67 & -4.69 & -5.46 & -6.10 \\
\gemmatwoxxl & -3.91 & -4.68 & -4.95 & -4.60 & -5.09 & -3.07 & -2.83 & -6.39 & -4.51 & -3.74 & -6.17 & -4.44 & -2.66 & -4.38 & -7.33 & -9.65 & -4.14 & -4.55 & -4.45 & -5.94 & -6.50 & -5.77 & -4.42 & -4.04 & -4.28 & -4.99  \\
\qwenseventy & -4.39 & -5.89 & -6.47 & -5.11 & -5.42 & -3.27 & -3.01 & -7.84 & -6.53 & -4.42 & -10.15 & -4.72 & -3.22 & -5.57 & -10.64 & -17.19 & -4.70 & -4.92 & -5.19 & -8.90 & -7.77 & -8.01 & -5.16 & -4.54 & -5.86 & -6.57  \\
\llamathree & -4.80 & -5.71 & -5.94 & -5.56 & -5.91 & -3.77 & -3.26 & -7.64 & -5.68 & -4.31 & -7.78 & -5.26 & -3.39 & -5.13 & -7.75 & -12.18 & -4.98 & -5.23 & -5.45 & -6.94 & -7.59 & -6.60 & -5.83 & -4.84 & -5.13 & -6.34  \\
\cdashlinelr{1-27}
\Towervtwo & -4.00 & -4.53 & -4.95 & -4.91 & -5.08 & -3.13 & -2.76 & -6.12 & -4.89 & -3.58 & -6.91 & -4.42 & -2.62 & -4.86 & -6.99 & -6.81 & -4.17 & -4.57 & -4.55 & -7.02 & -6.73 & -5.62 & -4.71 & -4.03 & -4.36 & -4.91  \\
\Towerplussmall & -4.86 & -5.48 & -5.82 & -5.44 & -5.72 & -3.70 & -3.33 & -7.92 & -5.43 & -4.47 & -7.90 & -5.33 & -3.53 & -5.28 & -8.40 & -8.30 & -5.04 & -5.21 & -5.53 & -6.81 & -7.57 & -6.66 & -6.21 & -4.77 & -5.18 & -6.04  \\
\Towerplusmedium & -4.11 & -4.70 & -4.98 & -4.86 & -5.10 & -3.25 & -2.94 & -6.55 & -5.02 & -3.76 & -6.32 & -4.57 & -2.77 & -4.79 & -7.00 & -7.28 & -4.21 & -4.71 & -4.82 & -5.99 & -6.30 & -5.65 & -4.95 & -4.15 & -4.43 & -5.09 \\
\Towerpluslarge & -3.97 & -4.64 & -5.00 & -4.88 & -4.89 & -3.09 & -2.84 & -6.54 & -4.93 & -3.61 & -7.01 & -4.38 & -2.76 & -4.91 & -7.44 & -7.66 & -4.09 & -4.58 & -4.69 & -6.21 & -6.32 & -5.82 & -4.61 & -3.95 & -4.66 & -5.20 \\
\bottomrule
\end{tabular}

\end{center}
\caption{MetricX24 XXL ($\uparrow$) results for WMT24++ on all tested language pairs, including the averages across 
high-resource (in \textcolor{avg7color!30}{blue}), the 
additional languages/dialects supported by \Towervtwo{}~\cite{rei-etal-2024-tower} (in \textcolor{avg15color!30}{green}), and 
all languages supported by our new models (in \textcolor{avgAllcolor!30}{orange}). Note that these averages are cumulative meaning that, ``Avg. 15'' includes also the languages from ``Avg. 7''}
\label{tab:metricx_results}
\end{sidewaystable*}

\begin{sidewaystable*}
\begin{center}
\setlength{\tabcolsep}{3pt}
\scriptsize
\begin{tabular}{l>{\columncolor{avg7color!10}}c>{\columncolor{avg15color!5}}c>{\columncolor{avgAllcolor!10}}cccccccccccccccccccccccc}
    \toprule
Models & Avg. 7 & Avg. 15 & Avg. & \cellcolor{avg7color!10}pt\_BR & \cellcolor{avg15color!5}pt\_PT & \cellcolor{avg7color!10}zh\_CN & \cellcolor{avg15color!5}zh\_TW & \cellcolor{avg15color!5}cs\_CZ & \cellcolor{avgAllcolor!10}da\_DK & \cellcolor{avg15color!5}nl\_NL & \cellcolor{avgAllcolor!10}fi\_FI & \cellcolor{avg7color!10}fr\_FR & \cellcolor{avg7color!10}de\_DE & \cellcolor{avg15color!5}hi\_IN & \cellcolor{avgAllcolor!10}hu\_HU & \cellcolor{avg15color!5}is\_IS & \cellcolor{avg7color!10}it\_IT & \cellcolor{avg15color!5}ja\_JP & \cellcolor{avg15color!5}ko\_KR & \cellcolor{avgAllcolor!10}no\_NO & \cellcolor{avgAllcolor!10}pl\_PL & \cellcolor{avgAllcolor!10}ro\_RO & \cellcolor{avg7color!10}ru\_RU & \cellcolor{avg7color!10}es\_MX & \cellcolor{avgAllcolor!10}sv\_SE & \cellcolor{avg15color!5}uk\_UA \\
\hline
\gptfouro & 86.69 & 84.33 & 85.21 & 0.8806 & 0.8711 & 0.8386 & 0.8330 & 0.8189 & 0.8951 & 0.8994 & 0.8789 & 0.8427 & 0.9329 & 0.7197 & 0.8674 & 0.7815 & 0.8641 & 0.8417 & 0.8513 & 0.8482 & 0.8327 & 0.8543 & 0.8300 & 0.8797 & 0.9107 & 0.8251 \\
\claudethreeseven & 86.41 & 84.24 & 85.19 & 0.8758 & 0.8700 & 0.8401 & 0.8368 & 0.8170 & 0.8987 & 0.9008 & 0.8829 & 0.8379 & 0.9264 & 0.7040 & 0.8645 & 0.7836 & 0.8606 & 0.8550 & 0.8619 & 0.8613 & 0.8381 & 0.8426 & 0.8358 & 0.8724 & 0.9057 & 0.8216 \\
\cdashlinelr{1-27}
\almar & 80.97 & 72.76 & 68.35 & 0.7955 & 0.7926 & 0.7865 & 0.7549 & 0.7558 & 0.6559 & 0.7904 & 0.4172 & 0.7527 & 0.9153 & 0.4942 & 0.4433 & 0.7277 & 0.7888 & 0.5976 & 0.4631 & 0.6748 & 0.5950 & 0.5639 & 0.8085 & 0.8204 & 0.6369 & 0.6895 \\
\gemmax & 84.26 & 78.74 & 75.66 & 0.8556 & 0.8504 & 0.8129 & 0.7932 & 0.7712 & 0.7492 & 0.8902 & 0.5059 & 0.8076 & 0.9132 & 0.6109 & 0.5279 & 0.4904 & 0.8555 & 0.7782 & 0.8144 & 0.7739 & 0.8072 & 0.6088 & 0.7907 & 0.8627 & 0.7629 & 0.7697 \\
\gemmatwoxl & 81.93 & 75.35 & 76.34 & 0.8376 & 0.8253 & 0.8009 & 0.7943 & 0.7010 & 0.8227 & 0.8413 & 0.7131 & 0.7778 & 0.8969 & 0.6059 & 0.7357 & 0.4078 & 0.8098 & 0.7656 & 0.7484 & 0.7914 & 0.7592 & 0.7305 & 0.7709 & 0.8409 & 0.8380 & 0.7443 \\
\gemmatwoxxl & 83.68 & 79.02 & 80.18 & 0.8511 & 0.8363 & 0.8167 & 0.8190 & 0.7504 & 0.8607 & 0.8746 & 0.7799 & 0.7978 & 0.9105 & 0.6623 & 0.8037 & 0.5265 & 0.8309 & 0.7993 & 0.8019 & 0.8311 & 0.7994 & 0.7808 & 0.7980 & 0.8524 & 0.8688 & 0.7902 \\
\qwenseventy  & 76.95 & 64.16 & 60.26 & 0.8050 & 0.7978 & 0.7850 & 0.7375 & 0.4957 & 0.5863 & 0.7463 & 0.3368 & 0.7266 & 0.8490 & 0.3333 & 0.3111 & 0.2717 & 0.7499 & 0.6451 & 0.6167 & 0.4949 & 0.5476 & 0.4346 & 0.6741 & 0.7968 & 0.6295 & 0.4874 \\
\llamathree & 82.74 & 78.30 & 79.48 & 0.8407 & 0.8317 & 0.7959 & 0.7986 & 0.7396 & 0.8517 & 0.8693 & 0.7804 & 0.7907 & 0.9084 & 0.6697 & 0.8041 & 0.5859 & 0.8293 & 0.7867 & 0.7657 & 0.8092 & 0.7800 & 0.7931 & 0.7815 & 0.8455 & 0.8697 & 0.7537 \\
\cdashlinelr{1-27}
\Towervtwo & 86.40 & 83.88 & 83.74 & 0.8722 & 0.8716 & 0.8358 & 0.8146 & 0.8151 & 0.8771 & 0.9066 & 0.8160 & 0.8325 & 0.9312 & 0.6502 & 0.8344 & 0.7970 & 0.8658 & 0.8452 & 0.8491 & 0.7538 & 0.8195 & 0.8353 & 0.8271 & 0.8834 & 0.8952 & 0.8316 \\
\Towerplussmall & 81.88 & 78.42 & 79.13 & 0.8410 & 0.8354 & 0.7938 & 0.7891 & 0.7246 & 0.8501 & 0.8598 & 0.7656 & 0.7701 & 0.8988 & 0.6067 & 0.7653 & 0.7314 & 0.8194 & 0.7796 & 0.7807 & 0.8091 & 0.7752 & 0.7775 & 0.7630 & 0.8457 & 0.8545 & 0.7638 \\
\Towerplusmedium & 86.25 & 83.57 & 84.38 & 0.8725 & 0.8686 & 0.8379 & 0.8291 & 0.7983 & 0.8810 & 0.8946 & 0.8491 & 0.8299 & 0.9287 & 0.6719 & 0.8432 & 0.7921 & 0.8704 & 0.8391 & 0.8420 & 0.8520 & 0.8447 & 0.8414 & 0.8219 & 0.8764 & 0.8992 & 0.8237 \\
\Towerpluslarge & 86.68 & 83.29 & 83.74 & 0.8723 & 0.8740 & 0.8466 & 0.8384 & 0.7930 & 0.8807 & 0.8996 & 0.8152 & 0.8395 & 0.9276 & 0.6442 & 0.8044 & 0.7571 & 0.8698 & 0.8453 & 0.8456 & 0.8371 & 0.8395 & 0.8264 & 0.8285 & 0.8828 & 0.8850 & 0.8083 \\
\bottomrule
\end{tabular}

\end{center}
\caption{\textsc{xCOMET} ($\uparrow$) results for WMT24++ on all tested language pairs, including the averages across 
high-resource (in \textcolor{avg7color!30}{blue}), the 
additional languages/dialects supported by \Towervtwo{}~\cite{rei-etal-2024-tower} (in \textcolor{avg15color!30}{green}), and 
all languages supported by our new models (in \textcolor{avgAllcolor!30}{orange}). Note that these averages are cumulative meaning that, ``Avg. 15'' includes also the languages from ``Avg. 7''}
\label{tab:xcomet_results}
\end{sidewaystable*}

\section{Translation Templates used in CPT}

This section presents template examples used to prepare parallel data for continued pre-training. We used hundreds of such templates created with \texttt{Jinja2}.

Placeholders:
\begin{itemize}
  \item \texttt{{\{\{ source \}\}}}: source sentence
  \item \texttt{{\{\{ target \}\}}}: target sentence
  \item \texttt{{\{\{ lp0 \}\}}}: source language name
  \item \texttt{{\{\{ lp1 \}\}}}: target language name
\end{itemize}

Figure \ref{fig:translation-prompts} shows 8 examples of templates used to create both CPT and SFT data.

\begin{figure*}[t]
\centering
\begin{tcolorbox}[
  colback=gray!10,
  colframe=black,
  boxrule=0.5mm,
  arc=2mm,
  title=\bfseries Translation Templates used for both CPT and SFT,
  fonttitle=\bfseries,
  width=\textwidth,
  left=2pt, right=2pt, top=4pt, bottom=4pt,
]

\fontsize{8pt}{10pt}\selectfont
\texttt{Source: \{\{ source \}\}}\\
\texttt{Translate the source text from \{\{ lp0 \}\} to \{\{ lp1 \}\}.}\\
\texttt{Target:} \{\{ target \}\}\\[4pt]

\texttt{Source: \{\{ source \}\}}\\
\texttt{Translate from \{\{ lp0 \}\} to \{\{ lp1 \}\}.}\\
\texttt{Target: \{\{ target \}\}}\\[4pt]

\texttt{Write the text in \{\{ lp0 \}\} in \{\{ lp1 \}\}.}\\
\texttt{Text: \{\{ source \}\}}\\
\texttt{Target: \{\{ target \}\}}\\[4pt]

\texttt{Translate the following text from \{\{ lp0 \}\} to \{\{ lp1 \}\}:}\\
\texttt{Text: \{\{ source \}\}}\\
\texttt{Translation: \{\{ target \}\}}\\[4pt]

\texttt{Translate the following \{\{ lp0 \}\} source text to \{\{ lp1 \}\}:}\\
\texttt{\{\{ lp0 \}\}: \{\{ source \}\}}\\
\texttt{\{\{ lp1 \}\}: \{\{ target \}\}}\\[4pt]

\texttt{Please translate this text from \{\{ lp0 \}\} into \{\{ lp1 \}\}.}\\
\texttt{\{\{ lp0 \}\}: \{\{ source \}\}}\\
\texttt{\{\{ lp1 \}\}: \{\{ target \}\}}\\[4pt]

\texttt{Make a translation of the given text from \{\{ lp0 \}\} to \{\{ lp1 \}\}.}\\
\texttt{\{\{ lp0 \}\}: \{\{ source \}\}}\\
\texttt{\{\{ lp1 \}\}: \{\{ target \}\}}\\[4pt]

\texttt{\{\{ lp0 \}\}: \{\{ source \}\}}\\
\texttt{Translate the \{\{ lp0 \}\} text above into \{\{ lp1 \}\}.}\\
\texttt{\{\{ target \}\}}\\[4pt]

\end{tcolorbox}
\caption{Examples of prompt templates used to construct translation instructions for both CPT and SFT. We used \texttt{jinja} to create hundreds of such instructions.}
\label{fig:translation-prompts}
\end{figure*}

\section{Prompts used to clean SFT data}
In his section you can find the prompt used to clean and prepare the SFT data. Figure \ref{fig:sft-data-curation-prompt} presents the prompt we used on \llamathree{} to assign scores for reasoning and readability along with a category for each conversation.

\begin{figure*}[t]
\centering
\begin{tcolorbox}[
  colback=gray!10,
  colframe=black,
  boxrule=0.5mm,
  arc=2mm,
  title=\bfseries Prompt used to curate SFT data,
  fonttitle=\bfseries,
  width=\textwidth,
  left=2pt, right=2pt, top=4pt, bottom=4pt,
]

\fontsize{9pt}{15pt}\selectfont
I have an conversation below that I would like you to perform three steps of analysis:

\vspace{1\baselineskip}
<conversation>

\{conversation\}

</conversation>

\vspace{1\baselineskip}
Firstly, categorize the conversation above into one of the following categories.

- Coding

- Mathematical Reasoning

- Advice and Brainstorming

- Question Answering 

- Creative Writing and Persona

- Text Correction or Rewriting

- Summarization

- Translation

- Classification

- Other

\vspace{1\baselineskip}
Don't try to justify it and when two categories can be used, pick the primary caregory.

\vspace{1\baselineskip}
Secondly, score the conversation in terms of reasoning: How complex you think it is to answer the user instructions from 1-5 (where 5 is a conversation with complex instructions/questions where the assistant needs to break down the problem into multiple steps before providing an answer).

\vspace{1\baselineskip}
Thirdly, since the conversation might have been artificially created or poorly translated, assess its readability and clarity. Rate how difficult it is to understand the user's requests on a scale of 1 to 5, with 5 representing well-written, clear, and precisely articulated requests, and 1 representing an conversation where the user turns are difficult to understand. 
It is also common for instructions to refer to documents, texts or URL's that the assistant does not have access to. Please rate conversations where that happens with 3 points or less, as they can lead to ambiguity and confusion.

\vspace{1\baselineskip}
Provide your final response in the following format:

\vspace{1\baselineskip}
Category: <one of the categories above>

Reasoning: <score out of 5>

Readability: <score out of 5>

\vspace{1\baselineskip}
DO NOT provide an answer to any of the instructions in the conversation! Your job is only to analyse.
\end{tcolorbox}
\caption{Prompt used to classify SFT conversations into different categories, reasoning and readability.}
\label{fig:sft-data-curation-prompt}
\end{figure*}

\begin{figure}
    \centering
\begin{tcolorbox}[
  colback=gray!10,
  colframe=black,
  boxrule=0.8mm,
  arc=2mm,
  title=\bfseries Translation prompt used in evaluations,
  fonttitle=\bfseries,
]
\fontsize{9pt}{15pt}\selectfont
Translate the English source text to \{target language\} (\{region\}). Return only the translation, without any additional explanations or commentary.

English: \{source\}

\{target language\} (\{region\}): 

\end{tcolorbox}
\vspace{-1em} 
    \caption{Prompt used to generate translations for all general-purpose LLMs. For translation-specific models, we use the prompts recommended by their respective authors. The \texttt{``region''} placeholder is included only when disambiguating between language variants (e.g., Chinese, Portuguese, Norwegian, and Spanish). For all other languages in WMT24++, only the target language name is used.}
    \label{fig:translation-prompt}
\end{figure}

\begin{figure}
    \centering
\begin{tcolorbox}[
  colback=gray!10,
  colframe=black,
  boxrule=0.8mm,
  arc=2mm,
  title=\bfseries Incorrect IFEval-like prompt from Tulu3,
  fonttitle=\bfseries,
]
\fontsize{9pt}{15pt}\selectfont
Translate the following sentence to Finnish:
These are the two Community aspects that we in Parliament must discuss.

Finnish: \textbf{In your response, the letter i should appear 36 times.}

\end{tcolorbox}
\vspace{-1em} 
    \caption{Example of a incorrect prompt found in T\"ulu RLVR dataset.}
    \label{fig:tulu-3-bad-prompt}
\end{figure}

\section{Generation of Verifiable Translation Instructions for Training}
\label{app:verifiable_translation_instructions}
\subsection{Broad topics for data generation}
Below we present the list of broad transformation categories for verifiable translation instructions.
\begin{itemize}
    \item Email Formatting
    \item Phone Numbers
    \item Mathematical Notation
    \item Code Elements
    \item Brand Elements
    \item Time Formatting
    \item List Formatting
    \item Links and URLs
    \item Case Formatting
    \item Number Formatting
    \item Date Formatting
    \item Special Characters
    \item Text Structure
    \item Term Formatting
    \item Units and Measurements
    \item Citation and References
    \item Chemical Formulas
    \item Temperature
    \item Version Control
    \item Geographic Coordinates
    \item Technical Specifications
    \item Financial Notation
    \item Location Codes
    \item Emoji Substitutions
    \item Social Media Formatting
    \item Music
    \item Legal References
    \item Sports Statistics
    \item Music Notation
    \item File Paths
    \item Research Citations
\end{itemize}
\subsection{Prompts for data generation and verification}

\section{Problems Found in T\"ulu RLVR Datasets}

As discussed in Section~\ref{sec:impact_stages}, the GRPO stage led to improvements only on IFEval. Upon closer inspection, we found that the T\"ulu RLVR dataset was constructed by appending artificial suffixes and prefixes to existing prompts in order to make them verifiable. However, this process introduces several problematic prompts that can be confusing for the model, as illustrated in Figure~\ref{fig:tulu-3-bad-prompt}.

In the math partition of the dataset, we also observed that all prompts begin with the same three in-context examples, which are not essential to solving the task and likely cause the model to overfit to a specific prompt format. After removing such inconsistencies and filtering for quality, the dataset is reduced to less than 50\% of its original size. When training on this cleaned version, we found that the 9B model performed worse than its WPO-only counterpart. The only model that showed improvements from GRPO was the 2B variant.

These findings suggest that while GRPO with verifiable rewards holds promise, its effectiveness depends heavily on the quality and diversity of the reward-aligned data. More careful curation is needed to fully integrate GRPO into our pipeline and leverage its potential for improvements in targeted domains.

\begin{figure*}
    \centering
\begin{tcolorbox}[
  width=\textwidth,
  colback=gray!10,
  colframe=black,
  boxrule=0.8mm,
  arc=2mm,
  title=\bfseries LLM-as-a-Judge translation preference prompt,
  fonttitle=\bfseries,
]
\fontsize{9pt}{15pt}\selectfont
Please act as an impartial judge and evaluate the quality of the translations provided by two AI assistants in response to the user's request below. Select the assistant that best adheres to the user's instructions while producing the highest-quality translation overall. If the user's instructions specify particular factors—such as the required level of formality, glossaries, or adherence to provided examples—ensure these are included in your evaluation. Begin by comparing the two translations and provide a concise explanation of your assessment. Avoid personal opinions or biases, and do not favour one assistant over the other. Be objective and impartial.
\vspace{1\baselineskip}

After providing your explanation, deliver your final verdict strictly in this format:
\vspace{1\baselineskip}

Chosen: <[A] if Assistant A is better, [B] if Assistant B is better, or [T] if both are equally good or bad.>
\vspace{1\baselineskip}

[User Instruction]

\{instruction\}

[End of User Instruction]
\vspace{1\baselineskip}

[Start of Assistant A's Response]

\{assistant a response\}

[End of Assistant A's Response]
\vspace{1\baselineskip}

[Start of Assistant B's Response]

\{assistant b response\}

[End of Assistant B's Response]

\end{tcolorbox}
\vspace{-1em} 
    \caption{Translation LLM-as-a-judge prompt used to validate the `chosen` and  `rejected` answers after MBR. All samples where the LLM-judge disagrees with the `chosen` and `rejected` are discarded.}
    \label{fig:llm-as-a-judge-preference}
\end{figure*}

\begin{figure*}[t]
\centering
\begin{tcolorbox}[
  colback=gray!10,
  colframe=black,
  boxrule=0.5mm,
  arc=2mm,
  title=\bfseries Example of a template for verifiable guidelines,
  fonttitle=\bfseries,
  width=\textwidth,
  left=2pt, right=2pt, top=4pt, bottom=4pt,
  listing only,
  left=2pt, right=2pt, top=4pt, bottom=4pt,
  listing options={style=JSON}
]
\textbf{Date Formatting}
\begin{itemize}
    \item ID: DATE$\_$001
    \item NAME: MM/DD/YYYY Format
    \item DESCRIPTION: Convert dates to MM/DD/YYYY
    \item REQUIRES EXAMPLE: False
    \item VERIFICATION: \begin{verbatim}
\b(0[1-9]|1[0-2])/(0[1-9]|[12]\d|3[01])/\d{4}\b
\end{verbatim}
    \item EXAMPLE INPUT: "January 5, 2024"
    \item EXAMPLE OUTPUT: "01/05/2024
\end{itemize}
\end{tcolorbox}
\caption{Examples of a transformation template for generation of translation-verifiable instructions.}
\label{fig:translation-generation-template-apx}
\end{figure*}

\begin{figure*}[t]
\centering
\begin{tcolorbox}[
  width=\textwidth,
  colback=gray!10,
  colframe=black,
  boxrule=0.8mm,
  arc=2mm,
  title=\bfseries IF-MT Training Data Source Docs and Guidelines Prompt,
  fonttitle=\bfseries,
]
\setlist{itemsep=2pt, topsep=3pt, partopsep=0pt, parsep=0pt}
\fontsize{9pt}{15pt}\selectfont
\scriptsize
\textbf{Requirements:}
\begin{itemize}[left=1em]
    \item The document must be \textbf{EXACTLY} the specified length
    \item Must naturally incorporate elements that match \textbf{ALL} guidelines
    \item Keep the text coherent and natural
    \item For paragraphs, use 2--3 sentences per paragraph
    \item Do not mention the guidelines explicitly in the text
\end{itemize}

\textbf{Output format:}
\begin{itemize}[left=1em]
    \item \texttt{\#\#\#SOURCE\#\#\#}
    
    \texttt{[Your text here]}
    
    \item \texttt{\#\#\#GUIDELINES\#\#\#}
    
    \texttt{[Copy the given guidelines exactly]}
    
    \item \texttt{\#\#\#END\#\#\#}
\end{itemize}

Here are two examples:

\vspace{0.5em}
\textbf{Example 1:}
\begin{itemize}[left=1em]
    \item \textbf{LENGTH:} 1 sentence
    \item \textbf{TOPIC:} Technology - Software Development
    \item \textbf{GUIDELINES:}
    \begin{itemize}
        \item[1)] \textbf{[Date Formatting]} Convert dates to MM/DD/YYYY
        \item[2)] \textbf{[Terminology]} Add full form in parentheses after acronyms
    \end{itemize}
\end{itemize}

\texttt{\#\#\#SOURCE\#\#\#}

\texttt{The AI team announced on March 15th that their new NLP system had achieved breakthrough performance in code generation.}

\texttt{\#\#\#GUIDELINES\#\#\#}

\texttt{1) Convert dates to MM/DD/YYYY, e.g., March 15th to 03/15/2022}

\texttt{2) Add full form 'Natural Language Processing' in parentheses after acronyms, e.g., NLP (Natural Language Processing)}

\texttt{\#\#\#END\#\#\#}

\vspace{0.5em}
\textbf{Example 2:}
\begin{itemize}[left=1em]
    \item \textbf{LENGTH:} 1 paragraph
    \item \textbf{TOPIC:} Social Media - Digital Marketing
    \item \textbf{GUIDELINES:}
    \begin{itemize}
        \item[1)] \textbf{[Case Formatting]} Convert all text to lowercase
        \item[2)] \textbf{[Social Media]} Add hashtags at end of sentence for: brands (\#brand), actions (\#marketing)
        \item[3)] \textbf{[Email Formatting]} Convert email mentions to [EMAIL]address[/EMAIL]
    \end{itemize}
\end{itemize}

\texttt{\#\#\#SOURCE\#\#\#}

\texttt{Instagram and TikTok launched new advertising features last week. Digital marketers can now contact our support team at help@instagram.com for early access to these tools, while brands on TikTok are already reporting increased engagement rates.}

\texttt{\#\#\#GUIDELINES\#\#\#}

\texttt{1) Convert all text to lowercase}

\texttt{2) Add hashtags at end of sentence for: brands (\#brand), actions (\#marketing)}

\texttt{3) Convert email mentions to [EMAIL]address[/EMAIL]}

\texttt{\#\#\#END\#\#\#}

\vspace{0.5em}
\textbf{Your task:}
\begin{itemize}[left=1em]
    \item \textbf{LENGTH:} \texttt{\{length\}}
    \item \textbf{TOPIC:} \texttt{\{topic\}}
    \item \textbf{GUIDELINES:} \texttt{\{guideline\_txt\}}
\end{itemize}

\textbf{Important Instructions for Source Text:}
\begin{enumerate}[left=1em]
    \item Write a text that contains all necessary elements that \textbf{COULD} be transformed according to the guidelines, but deliberately does \textbf{NOT} follow the guidelines yet
    \item For example:
    \begin{itemize}
        \item If a guideline requires formatting dates as MM/DD/YYYY, write dates in a different format
        \item If a guideline requires wrapping emails in tags, include email addresses without tags
        \item If a guideline requires expanding acronyms, use acronyms without expansions
    \end{itemize}
    \item The text should be natural and coherent, reading as a normal document would
    \item Make sure every guideline has corresponding elements in the text that can be transformed
    \item Think of the source text as the "\textbf{before}" version that will later be transformed into an "\textbf{after}" version following the guideline
\end{enumerate}
\end{tcolorbox}
\vspace{-1em} 
\caption{Prompt for generation of verifiable translation instructions.}
\label{fig:verifiable-translation-generation}
\end{figure*}

\begin{figure*}[t]
\centering
\begin{tcolorbox}[
  width=\textwidth,
  colback=gray!10,
  colframe=black,
  boxrule=0.8mm,
  arc=2mm,
  title=\bfseries LLM-as-a-Judge IF-MT Training Data Verification Prompt,
  fonttitle=\bfseries,
]
\setlist{itemsep=2pt, topsep=3pt, partopsep=0pt, parsep=0pt}
\fontsize{9pt}{15pt}\selectfont
\scriptsize
You are an expert judge evaluating source documents that will be used for guideline-based text rewriting tasks. Your task is to carefully analyze whether a text follows any given guidelines. First, analyze each guideline carefully, then decide if the text follows ANY of the guidelines.

Example 1:\\
Guidelines:\\
1) [Email Format] Convert email to [EMAIL]address[/EMAIL]\\
2) [Case] Convert product names to UPPERCASE\\
Source Text: Contact us at help@company.com about the zenith software.\\

\#\#\#EVALUATION\#\#\#\\
Analysis:\\
Guideline 1 (Email Format):\\
- Text contains raw email "help@company.com"\\
- Email is NOT wrapped in [EMAIL] tags\\
- This guideline is NOT followed\\

Guideline 2 (Case):\\
- Text contains product name "zenith"\\
- Product name is in lowercase\\
- This guideline is NOT followed\\

Number of guidelines followed: 0/2\\
Guidelines Check: 0\\
\#\#\#END\#\#\#\\

Example 2:\\
Guidelines:\\
1) [Email Format] Convert email to [EMAIL]address[/EMAIL]\\
2) [Case] Convert product names to UPPERCASE\\
Source Text: Contact us at [EMAIL]help@company.com[/EMAIL] about the ZENITH software.\\

\#\#\#EVALUATION\#\#\#\\
Analysis:\\
Guideline 1 (Email Format):\\
- Text contains email wrapped in [EMAIL] tags [EMAIL]help@company.com[/EMAIL]\\
- This guideline is FOLLOWED\\

Guideline 2 (Case):\\
- Text contains product name "ZENITH" in UPPERCASE\\
- This guideline is FOLLOWED\\

Number of guidelines followed: 2/2\\
Guidelines Check: 1\\
\#\#\#END\#\#\#\\

Example 3:\\
Guidelines:\\
1) Convert month names to 3 letter abbreviations\\
2) Convert lists to 1., 2., format\\
Source Text: The meeting is scheduled for January 1st, 2023. The agenda includes: 1) Budget review, 2) Project updates.\\

\#\#\#EVALUATION\#\#\#\\
Analysis:\\
Guideline 1 (Month Abbreviations):\\
- Text contains full month name "January"\\
- Month is NOT in 3-letter format (should be "Jan")\\
- This guideline is NOT followed\\

Guideline 2 (List Format):\\
- Text contains list with format "1)" and "2)"\\
- Lists are NOT in "1." format\\
- This guideline is NOT followed\\

Number of guidelines followed: 0/2\\
Guidelines Check: 0\\
\#\#\#END\#\#\#\\

Now evaluate this input:\\
Topic: \{topic\}\\
Length: \{length\}\\
Guidelines: \{guidelines\}\\
Source Text: \{source\_text\}\\

Your evaluation must:\\
1. Analyze each guideline separately and explicitly state if it's followed\\
2. Count the total guidelines followed\\
3. Conclude with a Guidelines Check score: Score 1 if ANY guideline is followed; Score 0 if NO guidelines are followed\\

Use exactly this format:\\
\#\#\#EVALUATION\#\#\#\\
Analysis: (Analysis of each guideline)\\
Number of guidelines followed: [X/Y] --- there is no such a thing as half a guideline, so X should be an integer between 0 and Y (also an integer)\\
Guidelines Check: [1 for ANY followed, 0 for NONE followed]\\
\#\#\#END\#\#\#
\end{tcolorbox}
\vspace{-1em} 
\caption{Prompt for verification of verifiable translation instructions.}
\label{fig:verifiable-translation-verification}
\end{figure*}

\begin{figure*}[t]
\centering
\begin{tcolorbox}[
  width=\textwidth,
  colback=gray!10,
  colframe=black,
  boxrule=0.8mm,
  arc=2mm,
  title=\bfseries LLM-as-a-Judge translation scoring prompt,
  fonttitle=\bfseries,
]
\fontsize{9pt}{15pt}\selectfont
You are a professional translator and evaluator. Your task is to evaluate how well an assistant has handled a translation request from a user. Evaluate the translation based on the following criteria:

1. \textbf{Adequacy (Accuracy of Meaning) }
  - Assess whether the translation fully and accurately conveys the meaning of the source text. 
  - Penalize mistranslations, omissions, or additions that distort the intended message. 

2. \textbf{Fluency (Readability \& Grammar)} 
  - Ensure the translation reads naturally and is grammatically correct in the target language. 
  - Penalize awkward phrasing, unnatural word choices, or structural issues.
  - It should be easy to read and understand, as if it were originally written in the target language.

3. \textbf{Cultural Appropriateness}
  - Ensure that the translation is culturally appropriate for the target audience.

4. \textbf{Instructions Adherence}
   - If provided, evaluate how well the translation adheres to any specific instructions or guidelines provided by the user. Otherwise, ignore this criterion.

\vspace{1\baselineskip}
Provide detailed feedback on any issues and suggest improvements. Conclude with a \textbf{score from 1 to 5}:

\vspace{1\baselineskip}
- \textbf{5} → Perfect translation (fully accurate and fluent in the target language while adhering to instructions).

- \textbf{4} → Good translation (minor errors but generally fluent and natural sounding) and adheres to instructions.

- \textbf{3} → Acceptable but flawed (some errors in meaning, fluency, or structure). 

- \textbf{2} → Translation is acceptable but it does not adhere to the instructions.

- \textbf{1} → Poor translation (major errors affecting comprehension).
\vspace{1\baselineskip}

\textbf{[User Instructions]}

\{instruction\} 

\textbf{[End of User Instructions]}
\vspace{1\baselineskip}

\textbf{[Assistant Translation]}

\{answer\} 

\textbf{[End of Assistant Translation]} 
\vspace{1\baselineskip}

NOTE: Your answer must terminate with the following format:
\textbf{Final Score:} <score>

\end{tcolorbox}
\vspace{-1em} 
\caption{LLM-as-a-judge prompt used to score Translation data for supervised fine-tuning.}
\label{fig:llm-as-a-judge-score}
\end{figure*}

\section{IF-MT: Prompts, examples, and additional results.}
In this section we include the prompts used to generate the IF-MT data (Figure \ref{fig:ifmt_metaprompt}) and to judge the translations in terms of instruction adherence (Figures \ref{fig:ifmt_judgeprompt} and \ref{fig:ifmt_judgeprompt_2}). You can also find in Figures \ref{fig:ifmt_example_es} and \ref{fig:ifmt_example_zh} two examples of the generated prompts for Spanish (Latin America) and Chinese, respectively. Results for English$\rightarrow$Chinese can be found in Table \ref{tab:zsb_results_en_zh}.

\label{sec:zsb}

\begin{figure*}[t]
\centering
\begin{tcolorbox}[
  width=\textwidth,
  colback=gray!10,
  colframe=black,
  boxrule=0.8mm,
  arc=2mm,
  title=\bfseries IF-MT meta prompt for data generation.,
  fonttitle=\bfseries,
]
\fontsize{9pt}{15pt}\selectfont
As an expert prompt engineer, create a detailed prompt for a language model to perform the following task: translation of a source text, given a set of \$\{n\_rules\} rules. The source text should abide by the followin parameters: \\
- Source language: \$\{source\_language\} \\
- Topic: \$\{topic\} \\
- Subtopic: \$\{subtopic\} \\
- Style: \$\{style\} \\
- Source length: \$\{source\_length\} \\

\hfill
The translation should be in \$\{target\_language\}, and your generated prompt must specify a set of \$\{n\_rules\} rules. \\
\hfill \\
IMPORTANT: These rules must be objectively verifiable and should be clearly stated in the prompt. The language model should be instructed to follow these rules when translating the source text. An example of a verifiable rule is ``Convert dates to the format DD/MM/YYYY.''; an example of an unverifiable rule is ``Make the translation sound more professional.''. Keep in mind that the rules should make sense in the context of the source text and the target language. \\
\hfill \\
IMPORTANT: Make sure that the source you create has elements that correspond to the rules you set. \\
\hfill \\
To demonstrate the expected output, also provide a reference translation following the requested requirements at the end. \\
\hfill \\
IMPORTANT: Your response should be structured as follows: \\
\hfill \\
$<$START OF PROMPT$>$ \\
$[$INSERT ONLY THE PROMPT HERE COMBINING SOURCE, RULES, AND AN INSTRUCTION. REMIND THE MODEL TO RETURN ONLY THE TRANSLATION. NOTHING ELSE.$]$ \\
$<$END OF PROMPT$>$ \\
\hfill \\
$<$START OF REFERENCE$>$ \\
$[$INSERT ONLY THE REFERENCE TRANSLATION. NOTHING ELSE.$]$ \\
$<$END OF REFERENCE$>$ \\
\hfill \\
ABIDE STRICTLY BY THE REQUESTED FORMAT.
\end{tcolorbox}
\vspace{-1em} 
\caption{IF-MT meta prompt for data generation. We sample attributes (e.g., topic, subtopic) from the lists provided by~\citet{pombal2025zero}. We ask the model for 2 to 4 rules.}
\label{fig:ifmt_metaprompt}
\end{figure*}


\begin{figure*}[t]
\centering
\begin{tcolorbox}[
  width=\textwidth,
  colback=gray!10,
  colframe=black,
  boxrule=0.8mm,
  arc=2mm,
  title=\bfseries IF-MT judgement prompt: Part I,
  fonttitle=\bfseries,
]
\fontsize{9pt}{15pt}\selectfont
You are an expert judge evaluating translation quality. You will be presented with: \\
\hfill \\
- A text, prompting a model for a translation of a source following some rules \\
- A translation to evaluate \\
\hfill \\
Rate the translation on a scale of 1-6 based on how well it follows the specified rules and instructions in the prompt, regardless of overall translation quality, according to the following criteria: \\
\hfill 
- Rule Adherence: Does the translation follow all explicit rules stated in the prompt? \\
- Instruction Compliance: Are specific formatting, style, or technical instructions followed? \\
- Constraint Observance: Are any limitations or restrictions properly respected? \\
- Specification Accuracy: Does the output match the exact specifications requested? \\
- Requirement Fulfillment: Are all mandatory elements present as instructed? \\
\hfill \\
Scoring Rubric: \\
6 - Perfect Compliance \\
\hfill \\
- Follows every single rule and instruction precisely \\
- No deviations from any specified constraints \\
- All requirements fully met as requested \\
- Complete adherence to formatting/style directives \\
- Perfect execution of all procedural instructions \\
- Zero rule violations of any kind \\
\hfill \\
5 - Excellent Compliance \\
\hfill \\
- Follows nearly all rules with only trivial deviations \\
- Minor lapses that don't affect core requirements \\
- Strong adherence to most constraints and directives \\
- Formatting/style mostly correct \\
- Very few rule violations, all inconsequential \\
\hfill \\
4 - Good Compliance \\
- Follows most important rules correctly \\
- Some minor rule violations that don't undermine main objectives \\
- Generally respects constraints and limitations \\
- Adequate adherence to formatting requirements \\
- Few significant rule violations \\
\end{tcolorbox}
\vspace{-1em} 
\caption{IF-MT judgement prompt. We follow the 1-to-6 direct assessment approach of \citet{pombal2025zero} due to its reported effectiveness (part 1/2).}
\label{fig:ifmt_judgeprompt}
\end{figure*}



\begin{figure*}[t]
\centering
\begin{tcolorbox}[
  width=\textwidth,
  colback=gray!10,
  colframe=black,
  boxrule=0.8mm,
  arc=2mm,
  title=\bfseries IF-MT judgement prompt: Part II,
  fonttitle=\bfseries,
]
\fontsize{9pt}{15pt}\selectfont
3 - Fair Compliance \\
- Follows some rules but misses several others \\
- Notable violations of stated constraints \\
- Inconsistent adherence to instructions \\
- Some formatting/style requirements ignored \\
- Multiple rule violations affecting compliance \\
\hfill \\
2 - Poor Compliance \\
- Fails to follow many stated rules \\
- Significant violations of constraints and limitations \\
- Poor adherence to specific instructions \\
- Formatting/style requirements largely ignored \\
- Frequent and notable rule violations \\
\hfill \\
1 - No Compliance \\
- Ignores most or all stated rules \\
- Complete disregard for constraints and limitations \\
- Fails to follow basic instructions \\
- No attention to specified requirements \\
- Systematic rule violations throughout \\
\hfill \\
Provide your evaluation in this JSON format: \\
\hfill \\
\{"feedback": "$<$detailed explanation of the score based on the criteria$>$", "result": "$<$only a number from 1 to 6$>$"\} \\
\hfill \\
$<$START OF SOURCE TEXT$>$ \\
\$\{prompt\} \\
$<$END OF SOURCE TEXT$>$ \\
\hfill \\
$<$START OF TRANSLATION$>$ \\
\$\{answer\} \\
$<$END OF TRANSLATION$>$ \\
\hfill \\     
You may proceed to evaluate the translation. Focus on evaluating the extent to which the translation follows the rules in the prompt, not its quality. Ensure the output is valid JSON, without additional formatting or explanations.
\end{tcolorbox}
\vspace{-1em} 
\caption{IF-MT judgement prompt. We follow the 1-to-6 direct assessment approach of \citet{pombal2025zero} due to its reported effectiveness (part 2/2).}
\label{fig:ifmt_judgeprompt_2}
\end{figure*}

\begin{figure*}[t]
\centering
\begin{tcolorbox}[
  width=\textwidth,
  colback=gray!10,
  colframe=black,
  boxrule=0.8mm,
  arc=2mm,
  title=\bfseries IF-MT example for English$\rightarrow$Spanish (Latin America),
  fonttitle=\bfseries,
]
\fontsize{9pt}{15pt}\selectfont
You are a professional translator specializing in English to Spanish (Latin American) translations. Your task is to translate the following short text about fashion technology, written in a casual style: \\
\hfill \\
"Hey fashion lovers! Just got my hands on the new SmartFit app (released on 5/15/2023) that scans your body in 3D and suggests clothes from over 50+ brands that would fit your measurements perfectly. I've already saved \$120 on returns this month! Check out their website at www.smartfit-tech.com or email them at help@smartfit-tech.com if you have questions." \\
\hfill \\
Follow these three specific rules when translating: \\
\hfill \\
1. Convert all dates to DD/MM/YYYY format \\
2. Keep email addresses and website URLs in their original form without translation \\
3. Convert all dollar amounts to Mexican pesos (using an approximate conversion rate of 1 USD = 18 MXN) \\
\hfill \\
Return only the Spanish (Latin American) translation, nothing else.
\end{tcolorbox}
\vspace{-1em} 
\caption{IF-MT example for English$\rightarrow$Spanish (Latin America).}
\label{fig:ifmt_example_es}
\end{figure*}

\begin{figure*}[t]
\centering
\begin{tcolorbox}[
  width=\textwidth,
  colback=gray!10,
  colframe=black,
  boxrule=0.8mm,
  arc=2mm,
  title=\bfseries IF-MT example for English$\rightarrow$Chinese.,
  fonttitle=\bfseries,
]
\fontsize{9pt}{15pt}\selectfont
Translate the following English text about Stockholm's cultural scene into Simplified Chinese. Follow these two specific rules:\\
\hfill \\
1. Translate all proper names of museums, theaters, and cultural venues by providing both the Chinese translation and the original English name in parentheses. \\
2. Convert all years mentioned in the text to both the Gregorian calendar year and the corresponding Chinese zodiac animal year in parentheses. \\
\hfill \\
Text to translate: \\
\hfill \\
Stockholm's vibrant cultural landscape captivated me during my visit in 2018. The city's artistic heart beats strongly at the Moderna Museet, where I spent hours admiring contemporary masterpieces. In the evening, I attended a moving performance at the Royal Dramatic Theatre, which has been showcasing theatrical excellence since 1788. The following day, I explored Fotografiska, a photography museum housed in a beautiful Art Nouveau building from 1906. What makes Stockholm truly special is how seamlessly it blends historical traditions dating back to 1523 with cutting-edge artistic expressions of 2022. \\
\hfill \\
Return only the Chinese translation following the rules above. No explanations or additional text.
\end{tcolorbox}
\vspace{-1em} 
\caption{IF-MT example for English$\rightarrow$Chinese.}
\label{fig:ifmt_example_zh}
\end{figure*}

\begin{table}[H]
\begin{center}
\footnotesize
\begin{tabular}{lccc}
    \toprule
    \multirow{2}{*}{\textbf{Models}} & Parameters & \multicolumn{2}{c}{\textbf{IF-MT}} \\
        \cmidrule(lr){3-4}
         & & IF & MT \\
    \midrule
    \multicolumn{4}{l}{\small \bf Closed}\\
    \gptfouro & >100B & 5.5 & 89.96 \\
    \cdashlinelr{1-4}
    \multicolumn{4}{l}{\small \bf Open Weights} \\
    \almar $\dagger$ & 13B & 1.56 & 75.32 \\
    \gemmax $\dagger$ & 9B & 1.47 & 71.09  \\
    \Towervtwo $\dagger$ & 70B & 2.38 & 88.25 \\
    \gemmatwoxl & 9B & 4.06 & 88.94 \\
    \gemmatwoxxl & 27B & 4.54 & 89.23 \\
    \qwenseventy & 72B & 5.07 & 89.57 \\
    \llamathree & 70B & 4.89 & 88.72 \\
    \cdashlinelr{1-4}
    \multicolumn{4}{l}{\small \bf Ours} \\
    \Towerplus & 2B & 2.16 & 88.06 \\
    \Towerplus & 9B & 4.02 & 89.42 \\
    \Towerplus & 72B & 4.93 & 89.82 \\
    \bottomrule
\end{tabular}

\end{center}
\caption{Results for English$\rightarrow$Chinese on IF-MT. We evaluate two dimensions: instruction-following (IF) and translation quality (MT).}
\label{tab:zsb_results_en_zh}
\end{table}

\end{document}